\documentclass[11pt]{article}
\usepackage[normalem]{ulem}

\usepackage[utf8]{inputenc} 
\usepackage[T1]{fontenc}    
\usepackage[numbers]{natbib}
\usepackage{hyperref}       
\usepackage{url}            
\usepackage{booktabs}       
\usepackage{amsfonts}       
\usepackage{nicefrac}       
\usepackage{microtype}      
\usepackage{xcolor}         

\usepackage{amsmath}
\usepackage{amssymb}
\usepackage{amsthm}

\usepackage{ stmaryrd }
\usepackage{algorithm}
\usepackage{algpseudocode}
\usepackage{cleveref}
\usepackage{graphicx}
\usepackage{caption}
\usepackage{subcaption}
\usepackage{enumitem}
\usepackage[subtle]{savetrees}

\newcommand{\R}{\mathbb{R}}
\newcommand{\G}{\mathcal{G}}
\newcommand{\C}{\mathcal{C}}
\newcommand{\p}{\mathcal{P}}
\newtheorem{theorem}{Theorem}
\newtheorem{prop}{Proposition}

\makeatletter
\newcommand{\vast}{\bBigg@{4}}
\newcommand{\Vast}{\bBigg@{5}}
\makeatother

\title{Tensor-Based Synchronization and the \\ Low-Rankness of the Block Trifocal Tensor}
\date{}
\author{ Daniel Miao \thanks{School of Mathematics, University of Minnesota (\href{mailto:miao0022@umn.edu}{miao0022@umn.edu}, \href{mailto:lerman@umn.edu}{lerman@umn.edu}) }
\hspace{1cm}
Gilad Lerman\footnotemark[1]
\hspace{1cm}
Joe Kileel\thanks{Department of Mathematics and Oden Institute for Computational Engineering and Sciences, University of Texas at Austin (\href{mailto:jkileel@math.utexas.edu}{jkileel@math.utexas.edu})}}

\begin{document}

\bibliographystyle{unsrt}

\maketitle

\begin{abstract}


The block tensor of trifocal tensors provides crucial geometric information on the three-view geometry of a scene. 
The underlying synchronization problem seeks to recover camera poses (locations and orientations up to a global transformation) from the block trifocal tensor.  
We establish an explicit Tucker factorization of this tensor, revealing a low multilinear rank of $(6,4,4)$ independent of the number of cameras under appropriate scaling conditions. We prove that this rank constraint provides sufficient information for camera recovery in the noiseless case. 
The constraint motivates a synchronization algorithm based on the higher-order singular value decomposition of the block trifocal tensor. Experimental comparisons with state-of-the-art global synchronization methods on real datasets demonstrate the potential of this algorithm for significantly improving location estimation accuracy.
Overall this work suggests that higher-order interactions in synchronization problems can be exploited to improve performance, beyond the usual pairwise-based approaches. 


\end{abstract}

\section{Introduction}\label{introduction}
Synchronization is crucial for the success of many data-intensive applications, including structure from motion, simultaneous localization and mapping (SLAM), and community detection. This problem involves estimating global states from relative measurements between states. While many studies have explored synchronization in different contexts using pairwise measurements, few have considered measurements between three or more states. In real-world scenarios, relying solely on pairwise measurements often fails to capture the full complexity of the system. For instance, in networked systems, interactions frequently occur among groups of nodes, necessitating approaches that can handle higher-order relationships. Extending synchronization to consider measurements between three or more states, however, increases computational complexity and requires sophisticated mathematical models. Addressing these challenges is vital for advancing various technological fields. For example, higher-order synchronization can improve the accuracy of 3D reconstructions in structure from motion by leveraging more complex geometric relationships. In SLAM, it enhances mapping and localization precision in dynamic environments by considering multi-robot interactions. Similarly, in social networks, it could lead to more accurate identification of tightly-knit groups. Developing efficient algorithms to handle higher-order measurements will open new research avenues and make systems more resilient and accurate.  

In this work, we focus on a specific instance of the synchronization problem within the context of structure from motion in 3D computer vision, where each state represents the orientation and location of a camera. Traditional approaches rely on relative measurements encoded by fundamental matrices, which describe the relative projective geometry between pairs of images. Instead, we consider higher-order relative measurements encoded in trifocal tensors, which capture the projective information between triplets of images. Trifocal tensors uniquely determine the geometry of three views, even in the collinear case  \cite{hartley2003multiple}, making them more favorable than triplets of fundamental matrices for synchronization. To understand the structure and properties of trifocal tensors in multi-view geometry, we carefully study the mathematical properties of the block tensor of trifocal tensors. We then use these theoretical insights to develop effective synchronization algorithms.

\textbf{Directly relevant previous works.}
In the structure from motion problem, synchronization has traditionally been done using incremental methods, such as Bundler \cite{snavely2006photo} and COLMAP \cite{schonberger2016structure}. These methods process images sequentially, gradually recovering camera poses. However, the order of image processing can impact reconstruction quality, as error may significantly accumulate. Bundle adjustment \cite{bundle_adjustment00}, which jointly optimizes camera parameters and 3D points, has been used to limit drifting but is computationally expensive.

Alternatively, global synchronization methods have been proposed. These methods process multiple images simultaneously, avoiding iterative procedures and offering more rigorous and robust solutions. Global methods generally optimize noisy and corrupted measurements by exploiting the structure of relative measurements and imposing constraints. Many global methods solve for orientation and location separately, using structures on $SO(3)$ and the set of locations. Solutions for retrieving camera poses from pairwise measurements have been developed for camera orientations \cite{Govindu2004_liealgebraic,hartley2013rotation,robust_rotation,ChatterjeeG13_rotation,shi2020message,Basri2012rotation}, camera locations \cite{1dsfm14,ozyesil2015robust,GoldsteinHLVS16_shapekick}, and both simultaneously \cite{Rosen2019SESync,Arrigoni2016SE3,Cucuringu2012SLAM,Briales2017SEd}. Some methods explore the structure on fundamental or essential matrices \cite{sengupta2017new,kasten2019algebraic,kasten2019gpsfm}.

Several attempts to extract information from trifocal tensors include works by: Leonardos et al. \cite{leonardos2015metric}, which parameterizes calibrated trifocal tensors with non-collinear pinhole cameras as a quotient Riemannian manifold and uses the manifold structure to estimate individual trifocal tensors robustly; 
Larsson et al. \cite{larsson2020calibration}, which proposes minimal solvers to determine calibrated radial trifocal tensors for use in an incremental pipeline, handling distorted images with constraints invariant to radial displacement; and Moulon et al. \cite{moulon2013global}, which introduces a structure from motion pipeline, retrieving global rotations via cleaning the estimation graph and solving a least squares problem, and solving for translations by estimating trifocal tensors individually by  linear programs. 
To our knowledge, no prior works develop a global pipeline where the synchronization operates directly on trifocal tensors.

\textbf{Contribution of this work.} The main contributions of this work are as follows:
\begin{itemize}
\item We establish an explicit Tucker factorization of the block trifocal tensor when its blocks are suitably scaled, demonstrating a low multilinear rank of $(6,4,4)$. Moreover, we prove that this rank constraint is sufficient to determine the scales and fully characterizes camera poses in the noiseless case. 
\item We develop a method for synchronizing trifocal tensors by enforcing this low rank constraint on the block tensor. We validate the effectiveness of our method through tests on several real datasets in structure from motion.
\end{itemize}


\section{Low-rankness of the block trifocal tensor}\label{sec2}
We first briefly review relevant background material in \Cref{sec:background}.  Then we present the main new construction and theoretical results in \Cref{low_rank_sec}.

\subsection{Background}\label{sec:background}
\subsubsection{Cameras and 3D geometry}
Given a collection of $n$ images $I_1,\dots,I_n$ of a 3D scene, let $t_i \in \R^3$ and $R_i\in SO(3)$ denote the location and orientation of the camera associated with the image $I_i$ in the global coordinate system. 
Moreover, each camera is associated with a calibration matrix $K_i$ that encodes the intrinsic parameters of a camera, including the focal length, the principal points, and the skew parameter. Then, the $3\times 4$ camera matrix has the following form, $P_i = K_i R_i [I_{3\times 3}, -t_i]$ and is defined up to nonzero scale.
Three-dimensional world points $X$ are represented as $\R^4$ vectors in homogeneous coordinates, and the projection of $X$ {into} the image corresponding to $P$ is $x = PX$.  3D world lines $L$ can be represented via Plücker coordinates as an $\mathbb{R}^6$ vector. Then the projection of $L$ onto the image corresponding to $P$ is $l = \p L$, where $\p$ is the $3\times 6$ line projection matrix. It can be written as  $\p = \left[ P^2 \wedge P^3; \, P^3 \wedge P^1; \, P^1 \wedge P^2\right]$ where $P^{i}$ is the $i$-th row of the camera matrix $P$ and wedge denotes exterior product.
Explicitly the $(i,j)$ element of the line projection matrix can be calculated as the determinant of the submatrix, where the $i$-th row is omitted and the column are selected as the $j$-th pair from $[(1,2), (1,3), (1,4), (2,3), (2,4), (3,4)]$. The elements on the second row are multiplied by $-1$.  

To retrieve global poses, relative measurement of pairs or triplets of images is needed. Let $x_i$ and $x_j$ be any pair of corresponding keypoints in images $I_i$ and $I_j$ respectively, meaning that they are images of a common world point. The fundamental matrix $F_{ij}$ is a $3\times 3$ matrix such that $x_i^TF_{ij} x_j = 0$. It is known that $F_{ij}$ encodes the relative orientation $R_{ij} = R_iR_j^T$ and translation $t_{ij} = R_i(t_i-t_j)$ through 
 $F_{ij} = K_i^{-T}[t_{ij}]_\times R_{ij}K_j^{-1}$. The essential matrix corresponds to the calibrated case, where $K_i = I_{3\times 3}$ for all $i$. 

\subsubsection{Trifocal tensors}

Analogous to the fundamental matrix, the trifocal tensor $T_{ijk}$ is a $3\times 3 \times 3$ tensor that relates the features across images and characterizes the relative pose between a triplet of cameras $P_i, P_j, P_k$. The trifocal tensor $T_{ijk}$ corresponding to cameras $P_i, P_j, P_k$ can be calculated by
\begin{align} \label{det def for tft}
    (T_{ijk})_{wqr} = (-1)^{w+1} \det\begin{bmatrix}
    \sim P_i^w\\
    P_j^q \\
    P_k^r
\end{bmatrix},
\end{align} 
where $P_i^w$ is the $w$-th row of $P_i$, and $\sim P_i^w$ is the $2\times 4$ submatrix of $P_i$ omitting the $w$-th row. The trifocal tensor determines the geometry of three cameras up to a global projective ambiguity, or up to a scaled rigid transformation in the calibrated case. In addition to point correspondences, trifocal tensors satisfy constraints for corresponding lines, and mixtures thereof. For example, let $l_i,l_j,l_k$ be corresponding image lines in the views of cameras $P_i,P_j,P_k$ respectively, then the lines are related through the trifocal tensor $T_{ijk}$ by $\left( l_j^T [(T_{ijk})_{1::}, (T_{ijk})_{2::}, (T_{ijk})_{3::}] l_k\right)[l]_\times = 0^T$, where $[l]_\times$ denotes the $ \times 3$ skew-symmetric matrix corresponding to cross product by $l$.  We refer to \cite{hartley2003multiple} for more details of the properties of a trifocal tensor.  We include the standard derivation of the trifocal tensor in  \Cref{derivation_tft}.

Since corresponding lines put constraints on the trifocal tensor, one advantage of incorporating trifocal tensors into structure from motion pipelines is that trifocal tensors can be estimated purely from line correspondences or a mixture of points and lines. Fundamental matrices can not be estimated directly from line correspondences, so the effectiveness of pairwise  methods for datasets where feature points are scarce is limited. Furthermore, trifocal tensors have the potential to improve location estimation. From pairwise measurements, one can only get the relative direction but not the scale and the location estimation in the pairwise setting is a ``notoriously difficult problem" (quoting from pages 316-317 of \cite{ozyecsil2017survey}). However, trifocal tensors encode the relative scales of the direction and can greatly simplify the location estimation procedure. 
We refer to several works on characterizing the complexity of minimal problems for individual trifocal tensors \cite{kileel2017minimal,duff2019plmp}, and on developing  methods for solving certain minimal problems \cite{9969132},\cite{nister2007minimal}, \cite{elqursh2011line}, \cite{kuang2013pose}, \cite{kukelova2017clever}, \cite{miraldo2018minimal}, \cite{kileel2018distortion}. We also refer to  \cite{kileel2022snapshot} for a survey paper on structure from motion, which discusses minimal problem solvers from the perspective of computational algebraic geometry.

\subsubsection{Tucker decomposition and the multilinear rank of tensors}

We review basic material on the Tucker decomposition and the multilinear rank of a tensor. We refer to \citep{kolda2009tensor} for more details while adopting its notation. 
Let $T \in \R^{I_1 \times I_2\times \cdots \times I_N}$ be an order $N$ tensor. 
The \textit{mode-$i$} \textit{flattening} (or \textit{matricization}) $T_{(i)} \in \mathbb{R}^{I_i \times (I_1 \ldots I_{i-1} I_{i+1} \ldots I_{N})}$ is the rearrangement of $T$ into a matrix by taking mode-$i$ fibers to be columns of the flattened matrix. By convention, the ordering of the columns in the flattening follows lexicographic order of the modes excluding $i$. Symbols $\otimes$ and $\odot$ denote the Kronecker product and the Hadamard product respectively. The norm on tensors is defined as $\| T \| = \|T_{(1)}\|_F$. 
The $i$-rank of $T$ is the column rank of $T_{(i)}$ and is denoted as rank$_i(T)$. Let $R_i$=rank$_i(T)$. Then the \textit{multilinear rank} of $T$ is defined as mlrank($T$) = $(R_1,R_2,\dots,R_N)$. The \textit{$i$-mode product} of $T$ with a matrix $U\in \R^{m \times I_i}$ is a tensor in $\R^{I_1 \times \cdots \times I_{i-1} \times m \times I_{i+1} \times \cdots \times I_N}$ such that 
\begin{align*}
    (T\times_i U)_{j_1\cdots j_{i-1}kj_{i+1}\cdots j_N} = \sum_{j_i=1}^{I_i}T_{j_1j_2\cdots j_N} U_{kj_i}.
\end{align*} 
Then, the \textit{Tucker decomposition} of $T \in \R^{I_1 \times I_2\times \cdots \times I_N}$ is a decomposition of the following form:
\begin{align*}
    T = \G \times_1 A_1 \times_2 A_2 \times_3 \cdots \times_N A_N = \llbracket \G; A_1, A_2, \dots, A_N \rrbracket,
\end{align*}
where $\G \in \R^{Q_1 \times \cdots \times Q_N}$ is the core tensor, and $A_n \in \R^{I_n \times Q_n}$ are the factor matrices. Without loss of generality, the factor matrices { can be} assumed to have orthonormal columns. 
Given the multilinear rank of the core tensor $(R_1,\dots,R_N)$, the Tucker decomposition approximation problem can be written as
\begin{equation}\label{eq2}
    \underset{\G\in \R^{R_1\times \cdots \times R_N}, A_i \in \R^{I_i \times R_i}}{\arg \min} \| T - \llbracket \G; A_1, A_2, \dots, A_N \rrbracket \|.
\end{equation}
A standard way of solving \eqref{eq2} is the \textit{higher-order singular value decomposition} (\textit{HOSVD}). The HOSVD is computed with the following steps. First, for each $i$ calculate the factor matrix  $A_i$ as the $R_i$ leading left singular vectors of $T_{(i)}$. Second,  set the core tensor  $\G$  as  $\G = T \times_1 A_1^T \times_2  \cdots \times_N A_N^T$.
Though the solution from HOSVD will not be the optimal solution to \eqref{eq2}, it  satisfies a quasi-optimality property: if $T^*$ is the optimal solution, and $T'$ the solution from HOSVD, then 
{
\begin{equation}\label{quasi_optimality}
    \| T - T'\| \leq \sqrt{N} \|T - T^*\|.
\end{equation} }


\subsection{Low Tucker rank of the block trifocal tensor and one shot camera retrieval}\label{low_rank_sec}

Suppose we are given a set of camera matrices $\{P_i\}_{i=1}^n$ with $n\geq 3$ and scales fixed on each camera matrix. Define the \textit{block trifocal tensor} $T^n$ to be the $3n \times 3n \times 3n$ tensor, where the $3\times 3 \times 3$ sized $ijk$ block is the trifocal tensor corresponding to the triplet of cameras $P_i,P_j,P_k$. 
We assume for all blocks that have overlapping indices, the corresponding $3\times 3 \times 3$ tensor is also calculated using the formula \eqref{det def for tft}.  We summarize key properties of $T^n$ in \Cref{prop1} and  \Cref{theorem1}. The proof of Proposition~\ref{prop1} is by direct computation and can be found in \Cref{prop1proof}.
\begin{prop}\label{prop1}
 We have the following observations for the block trifocal tensor $T^n$. For all distinct $i, j \in [n]$, { we have the following properties:} 
\begin{itemize}
    \item[(i)] $T^n_{iii} = 0_{3\times 3 \times 3}$
    \item[(ii)] The $T^n_{jii}$ blocks are rearrangements of elements in the fundamental matrix $F_{ij}$ up to signs. 
    {\item[(iii)] The $T^n_{iji}$ and $T^n_{iij}$ blocks encode the epipoles.}
    \item[(iv)] The horizontal slices $T^n(i,:,:)$ of $T^n$ are skew symmetric. 
    \item[(v)]  {When all cameras are calibrated, three singular values of $T^n_{(1)}$ are equal.}
\end{itemize}
\end{prop}
\begin{theorem}[Tucker factorization and low multilinear rank of block trifocal tensor]\label{theorem1}
The block trifocal tensor $T^n$ admits a Tucker factorization, $T^n = \G \times_1 \p \times_2 \C \times_3 \C$, where $\G \in \R^{6\times 4 \times 4}$, $\p \in \R^{3n\times 6}$, and $\C \in \R^{3n\times 4}$. If the $n$ cameras that produce $T^n$ are not all collinear, then $mlrank(T^n) = (6,4,4)$. If the $n$ cameras that produce $T^n$ are collinear, then $mlrank(T^n) \preceq (6,4,4)$. 
\end{theorem}

\begin{proof}
    We can explicitly calculate that $T^n = \G \times_1 \p \times_2 \C \times_3 \C$. The details of the calculation are in  \Cref{thrm1proof}. The specific forms for $\G, \C,  \p$ are the following.  The horizontal slices of the core are
   \begin{tiny}  
\begin{align*}\label{core format}
    \vast\{ &\begin{bmatrix}
0 & 0 & 0 & 0\\
0 & 0 & 0 & 0\\
0 & 0 & 0 & 1\\
0 & 0 & -1 & 0
\end{bmatrix},
\begin{bmatrix}
0 & 0 & 0 & 0\\
0 & 0 & 0 & -1\\
0 & 0 & 0 & 0\\
0 & 1 & 0 & 0
\end{bmatrix}, \begin{bmatrix}
0 & 0 & 0 & 0\\
0 & 0 & 1 & 0\\
0 & -1 & 0 & 0\\
0 & 0 & 0 & 0
\end{bmatrix}, 
\begin{bmatrix}
0 & 0 & 0 & 1\\
0 & 0 & 0 & 0\\
0 & 0 & 0 & 0\\
-1 & 0 & 0 & 0
\end{bmatrix}, \begin{bmatrix}
0 & 0 & -1 & 0\\
0 & 0 & 0 & 0\\
1 & 0 & 0 & 0\\
0 & 0 & 0 & 0
\end{bmatrix}, 
\begin{bmatrix}
0 & 1 & 0 & 0\\
-1 & 0 & 0 & 0\\
0 & 0 & 0 & 0\\
0 & 0 & 0 & 0
\end{bmatrix} \vast\}. 
\end{align*} 
\end{tiny}
\!\!\! The factor matrices are
$\C = \begin{bmatrix}
    P_1, P_2, \dots, P_n
\end{bmatrix}^T\in \R^{3n\times 4}$ and $\p = \begin{bmatrix}
    S_1, S_2, \dots, S_n
\end{bmatrix}^T \in \R^{3n\times 6}$, where $P_i$ are the camera matrices and $S_i$ are the corresponding line projection matrices. 

Now, we suppose that the $n$ cameras are not collinear. We first show that $\C$ and $\p$ both have full rank. From \citep{hartley2003multiple}, the null space of a camera matrix $P_i$ is generated by the camera center. For the sake of contradiction, suppose that rank($\C$) < 4. Then there exists $x\in \R^4$ such that $x\not = 0 $ and $\C x = 0$. This means that $P_i x = 0$ for all $i=1,...,n$. Then, $x$ is the camera centre for all cameras, which means that the cameras are centered at one point and are collinear. Similarly, every vector in the null space of the line projection matrix $S_i$ is a line that passes through the camera centre \cite{hartley2003multiple}. For the sake of contradiction, suppose that rank($\p$) < 6. Then there exists $x\in \R^6$ such that $x\not = 0 $ and $\p x = 0$. This implies that $S_i x = 0$ for all $i=1,...,n$, which means that $x$ is a line that passes through all of the camera centers. Again the cameras are collinear, which is a contradiction.
Next we write the flattening of the block trifocal tensor as $T^n_{(1)} = \p \G_{(1)} (\C \otimes \C)^T$. 
Then $\p \in \R^{3n\times 6}$ has rank 6, and $(\C \otimes \C)^T\in \R^{16 \times 9n^2}$ has rank 16. Given the specific form of $\G$, where $\G_{(1)} \in \R^{6\times 16}$ it is easy to check rank($\G_{(1)}$) = 6. Thus, rank($T^n_{(1)}$) = 6. Similarly, we can show that rank($T^n_{(2)}$) = 4, and rank($T^n_{(3)}$) = 4. This implies that the multilinear rank of the block trifocal tensor is $(6,4,4)$ when the $n$ cameras are not collinear. 

When the $n$ cameras are collinear, the individual factors in each flattening may be rank deficient, so that rank($T^n_{(1)}$) $\leq 6$, rank($T^n_{(2)}$) $\leq 4$, and rank($T^n_{(3)}$) $\leq 4$.  This  implies mlrank($T^n$) $ \preceq (6,4,4)$.   
\end{proof}
The theorem inspires a straightforward way of retrieving global poses from the block trifocal tensor, which we summarize in the following claim.

\begin{prop}[One shot camera pose retrieval]\label{oneshot} Given the block trifocal tensor $T^n$ produced by cameras $P_1,P_2,...,P_n$, the cameras can be retrieved from $T^n$ up to a global projective ambiguity using the higher-order SVD. The cameras will be the leading $4$ singular vectors of $T^n_{(2)}$ or $T^n_{(3)}$. 

\end{prop}

Using the higher-order SVD on $T^n$, we can get a Tucker decomposition of the block trifocal tensor $T^n = \hat{\G} \times_1 \hat{\p} \times_2 \hat{\C} \times_3 \hat{\C}'$. Though the Tucker factorization is not unique \cite{kolda2009tensor}, as we can apply an invertible linear transformation to one of the factor matrices and apply the inverse onto the core tensor, this invertible linear transformation  {can be interpreted as the global projective ambiguity for projective} 3D reconstruction algorithms. Thus, the cameras { can be retrieved by taking the} leading four singular vectors of the mode-2 and mode-3 flattenings of the block tensor.

Very importantly however, in practice each trifocal tensor block in $T^n$ can be estimated from image data \textit{only up to an unknown multiplicative scale} \cite{hartley2003multiple}.  The following theorem establishes the fact that the multilinear rank constraints provide sufficient information for determining the correct scales.  In the statement $\odot_b$ denotes blockwise scalar multiplication, thus the $(i,j,k)$-block of $\lambda \odot_b T^n$ is $\lambda_{ijk} T^n_{ijk} \in \mathbb{R}^{3 \times 3 \times 3}$.

\begin{theorem}\label{thm:2}
Let $T^n \in \R^{3n \times 3n \times 3n}$ be a block trifocal tensor corresponding to $n \geq 4$ calibrated or uncalibrated cameras in generic position.  Let $\lambda \in \R^{n\times n \times n}$ be a block scaling with $\lambda_{ijk}$ nonzero iff $i,j,k$ are not all equal.  Assume that $\lambda \odot_b T^n \in \mathbb{R}^{3n \times 3n \times 3n}$ has multilinear rank $(6,4,4)$ where $\odot_b$ denotes blockwise scalar multiplication. 
Then there exist $\alpha, \beta, \gamma \in \R^n$ such that $\lambda_{ijk} = \alpha_i \beta_j \gamma_k$ whenever $i,j,k$ are not all  the  same.  
\end{theorem}

\begin{proof}[Sketch]
  The idea is to identify certain submatrices in the flattenings of $\lambda \odot_b T^n $ which must have determinant $0$, and use these  to solve for $\lambda$.   A proof is in Appendix \ref{app:proof2}. We remark that the proof technique extends that of \cite[Theorem~5.1]{muller2024commonlines}, which showed a similar result for a matrix problem. 
\end{proof}

\Cref{thm:2} is the basic guarantee for our algorithm development below. 
We stress that the ambiguities brought by $\alpha, \beta, \gamma$ are \textit{not} problematic for purposes of recovering the camera matrices by \Cref{oneshot}.  Indeed, $(\alpha \otimes \beta \otimes \gamma) \odot_b T^n = \mathcal{G} \times_1 (D_{\alpha} \mathcal{P}) \times_2 (D_{\beta} \mathcal{C}) \times_3 (D_{\gamma} \mathcal{C})$ where $D_{\alpha} \in \mathbb{R}^{3n \times 3n}$ is the diagonal matrix with each entry of $\alpha$ triplicated, etc.  Hence the camera matrices can still be recovered up to individual scales (as expected) and a global projective transformation, from the higher-order SVD.

\section{Synchronization of the block trifocal tensor}\label{sync_sec}
In this section, we develop a heuristic method for synchronizing the block trifocal tensor $T^n$ by exploiting the multilinear rank of $T^n$ from \Cref{theorem1}. Let $\hat{T}^n$ denote the estimated block trifocal tensor, and $T^n$ the ground truth. Assume that there are $n$ images and a set of trifocal tensor estimates $\Hat{T}_{ijk}$ where $(i,j,k)\in \Omega$ and $\Omega$ is the set of indices whose corresponding trifocal tensor is estimated. Note that each estimated trifocal tensor $\Hat{T}_{ijk}$ will have an unknown scale $\lambda_{ijk} \in \mathbb{R}^*$ associated with it. We always assume that we observe the $iii$ blocks, as they will be $0$. We formulate the block trifocal tensor $\hat{T}^n$ by plugging in the estimates $\Hat{T}_{ijk}$ and setting the unobserved positions ($(i,j,k) \not \in \Omega$) to $3\times 3 \times 3$ tensors of all zeros. Let $W_\Omega\in \{0,1\}^{3n\times 3n\times 3n}$ denote the block tensor where the $(i,j,k)$ blocks are ones for $(i,j,k)\in \Omega$ and zeros otherwise. Let $W_{\Omega^C}$ denote the opposite. 
In our experiments, we observe that the HOSVD is quite robust against noise for retrieving camera poses, which arises e.g., from numerical sensitivities when first estimating relative poses \cite{fan2022instability}.  Therefore we develop an algorithm that projects $\hat{T}^n$ onto the set of tensors that have multilinear rank of $(6,4,4)$ while completing the tensor and retrieving an appropriate set of scales. Specifically, we can write our problem as
\begin{equation}\label{opt_prob}
    { \underset{\Lambda}{\min} \| \Lambda \odot \Hat{T}^n - \mathcal{P}_\tau (\Lambda \odot \Hat{T}^n) \|^2}
\end{equation}
where $\Lambda \in \R^{3n \times 3n \times 3n}$, each $3\times 3 \times 3$ block is uniform, $\Lambda_{ijk}$ blocks are zero for $(i,j,k)\not \in \Omega$, and $\Lambda$ satisfies a normalization condition like $\| \Lambda \|^2 = 1$ to avoid its vanishing. However, we drop this normalization constant in our implementation as we never observe $\Lambda$ vanishing in practice. (For convenience, we formulate this section with the notation of $\Lambda \in \mathbb{R}^{3n \times 3n \times 3n}$ and Hadamard multiplication, rather than $\lambda \in \mathbb{R}^{n \times n \times n}$ and blockwise scalar multiplication from \Cref{thm:2}.)  
Furthermore in problem \eqref{opt_prob},  $\mathcal{P}_\tau$ denotes the exact projection onto the set $\Gamma = \{T \in \R^{3n\times 3n \times 3n} : \text{mlrank}(T) = (6,4,4)\}$. 
Note that though HOSVD provides an efficient way to project onto $\Gamma$, it is quasi-optimal and not the exact projection. The exact projection is much harder to calculate, and in general NP-hard.  The algorithm below adopts an alternating projection strategy to estimate the best set of scales. 




\subsection{Higher-order SVD with a hard threshold (HOSVD-HT)}
The key idea for our algorithm is to use the relative scales on the rank truncated tensor as a heuristic to retrieve scales for the estimated block tensor. There are two main challenges for calculating the rank truncated tensor. First, the exact projection $\p_\tau$ 
onto $\Gamma$ is expensive and difficult to calculate. Second, many blocks in the block tensor will be unknown if the corresponding images of the block lacks corresponding point and directly projecting the uncompleted tensor will be inaccurate. 
We apply an HOSVD framework with imputations to tackle the challenges. Regarding the first challenge, HOSVD is a simple, efficient, and quasi-optimal \eqref{quasi_optimality} projection onto $\Gamma$. Though inexact, it is a reliable approximation. For the second challenge, the tensor $\hat{T}^n$ must be completed. We adopt the matrix completion idea of HARD-IMPUTE \cite{mazumder2010spectral}, where the matrix is filled-in iteratively with the rank truncated matrix obtained using the hard-thresholded SVD. { In other words, w}e complete the missing blocks with the corresponding blocks in the rank truncated tensor. 
We define three hyperparameters $l_1,l_2,l_3$ that correspond to the thresholding parameters of the hard-thresholded SVD on modes $1,2,3$ of the block tensor respectively. Specifically, for each mode-$i$ flattening $T^n_{(i)}$, we calculate the full SVD $T^n_{(i)} = USV^T$. Since our tensor will scale cubically with the number of cameras, we suggest using a randomized SVD. We refer to \cite{halko2011finding} for different randomized strategies.  Assume the singular values $\sigma_i$ on the diagonal of $S$ are sorted in descending order, as usual. We return the factor matrix $A_i$ as the top $a$ left singular vectors in $U$, where $a = \max\{i:S_{ii} > l_i\}$. 
Our adapted truncation method is summarized by \Cref{alg:hosvdhi}.

\begin{algorithm}[htp!]
    \caption{HOSVD-HT}\label{alg:hosvdhi}
    \begin{algorithmic}
        \State \textbf{Input: } 
        $\hat{T}^n \in \R^{3n\times 3n \times 3n}$: the estimated block tensor; 
        $l_1,l_2,l_3 \in \R$: the thresholds for modes $1,2,3$ respectively
        \State \textbf{Output: } $\hat{T}_r \in \R^{3n\times 3n\times 3n}$: the rank truncated tensor.
        \For{i = 1 to 3}
        \State Perform the randomized SVD on the mode-$i$ flattening such that  $\hat{T}^n_{(i)} \gets U S {V^T}$
        \State $a_i\gets \max\{i: S_{ii} > l_i\}$
        \State $A_i \gets$ first $a_i$ columns of $U$ 
        \EndFor
        \State $\G = \hat{T}^n \times_1 A_1^T \times_2 A_2^T \times_3 A_3^T$
        \State $\Hat{T}^r\gets \llbracket \G; A_1, A_2, A_3 \rrbracket$
    \end{algorithmic}
\end{algorithm}

From now on, we refer to hard-thresholded HOSVD as HOSVD-HT and denote the operation as $\p_{ht}$. 

\subsection{Scale recovery}
HOSVD-HT provides an efficient way for projecting $\hat{T}^n$ onto the set of tensors with { with truncated rank}. To recover scales, we use the rank truncated tensor's relative scale as a heuristic to adjust the scale on our estimated block trifocal tensor $\hat{T}^{(n)}$. For each step, we solve \begin{align}\label{opt_prob2}
    \Lambda^{(t+1)} &= \underset{\Lambda}{\arg \min} \| \Lambda \odot \Hat{T}^n - \mathcal{P}_{ht} (\Lambda^{(t)} \odot \Hat{T}^n) \|^2 \quad \text{s.t. }\Lambda_{ijk} = 0_{3\times 3\times 3} \text{ for } (i,j,k) \in \Omega^C,
\end{align}
where we drop the normalization condition on $\Lambda$ because in practice it is not needed.
We solve \eqref{opt_prob2} for each observed block separately. Denoting $\mathcal{P}_{ht} (\Lambda^{(t)} \odot \Hat{T}^n)$ as $(\Hat{T}^n_r)^{(t)}$, {we have} 
\begin{align}\label{opt_prob3}
    \Lambda_{ijk}^{(t+1)} &= \underset{\mu}{\arg \min} \| \mu \cdot \Hat{T}_{ijk}^n - (\Hat{T}_r^n)_{ijk}^{(t)}\|^2 = \frac{trace((\Hat{T}_{ijk}^n)_{(1)}^T ((\Hat{T}^n_r)_{ijk}^{(t)})_{(1)})}{\|((\Hat{T}^n_r)_{ijk}^{(t)})_{(1)}\|_F^2}.
\end{align}

Recall that our strategy for completing the tensor is to impute the tensor with the entries from the rank truncated tensor using HOSVD-HT. Specifically, given the current imputed tensor $(\hat{T}^n)^{(t)}$, we calculate $\mathcal{P}_{ht}((\hat{T}^n)^{(t)})$ and the new scales $\Lambda^{(t+1)}$. Then update with
\begin{equation}\label{last_update}
    (\hat{T}^n)^{(t+1)} = (\Lambda^{(t+1)} \odot (\hat{T}^n)^{(t)} \odot W_\Omega) + \mathcal{P}_{ht}((\hat{T}^n)^{(t)}) \odot W_{\Omega^C}.
\end{equation}

\subsection{Synchronization algorithm}
Now we summarize our synchronization framework in \Cref{alg:sync}.
\begin{algorithm}[htp!]
    \caption{Synchronization of the block trifocal tensor}\label{alg:sync}
    \begin{algorithmic}
        \State \textbf{Input: } $\hat{T}^n \in \R^{3n\times 3n \times 3n}$; $W_\Omega, W_{\Omega^C} \in \{0,1\}^{3n\times 3n\times 3n}$; $l_1,l_2,l_3 \in \R$ 
        \State \textbf{Output: } $\C \in \R^{3n\times 4}$: camera matrices up to a $4\times 4$ projective ambiguity and camera-wise scales
        \State Initialize $\hat{T}^n$ by imputing unobserved blocks randomly to get $(\hat{T}^n)^{(0)}$
        \While{not converged}
            \State Calculate $\mathcal{P}_{ht}((\hat{T}^n)^{(t)})$ using HOSVD-HT
            \State Calculate $\Lambda^{(t+1)}$ \eqref{opt_prob2} using \eqref{opt_prob3}
            \State  $(\hat{T}^n)^{(t+1)} \gets (\Lambda^{(t+1)} \odot (\hat{T}^n)^{(t)} \odot W_\Omega) + \mathcal{P}_{ht}((\hat{T}^n)^{(t)}) \odot W_{\Omega^C}$
            \State $t\gets t+1$
        \EndWhile
        \State $(\G, A_1, A_2, A_3) \gets HOSVD((\hat{T}^n)^{(t)})$
        \State $\C \gets$ First 4 columns of $A_2$
    \end{algorithmic}
\end{algorithm}
We have observed that the algorithm can overfit, as the recovered scales will experience sudden and huge leaps. Our stopping criteria for the algorithm is when we observe sudden jumps in the variance of the new scales or when we exceed a maximum number of iterations. Another challenge in structure from motion datasets is that estimations may be highly corrupted. The HOSVD framework mainly consists of retrieving a dominant subspace from each flattening. Thus, it is natural to replace the SVD on each flattening with a more robust subspace recovery method, such as the Tyler's M estimator (TME) \cite{tyler1987distribution} or a recent extension of TME that incorporates the information of the dimension of the subspace in the algorithm \cite{yu2024subspace}. We refer to 
Appendix \ref{horste} for more details and provide an implementation there.

\section{Numerical experiments} \label{num_sec}

We conduct experiments of \Cref{alg:sync} on two benchmark real datasets, the EPFL datasets \cite{strecha2008benchmarking} and the Photo Tourism datasets \cite{1dsfm14}. 
We observe that the algorithm performs better in the calibrated setting, and since the calibration matrix is usually known in practice, we restrict our scope of experiments to calibrated trifocal tensors. We compare against { three} state-of-the-art synchronization based on two view measurements, NRFM \cite{sengupta2017new} and LUD \cite{ozyesil2015robust}. 
NRFM 
relies on nonconvex optimization and requires a good initialization. We test NRFM with an initialization obtained from LUD and with a random initialization. { We also test BATA \cite{Zhuang_2018_CVPR} initialized with MPLS \cite{shi2020message}. }
We refer to \ref{comprehensive_results} in the appendix for a comprehensive summary of numerical results including rotation and translation estimation errors. We include our code in the following github repository: \href{https://github.com/dmiao153/TrifocalSync}{TrifocalSync}. 


\subsection{EPFL dataset}
For EPFL, we follow the experimental setup and adopt code from \cite{julia2018critical} and test an entire structure from motion pipeline. We first describe the structure from motion pipeline for EPFL experiements.
\begin{itemize}[wide=0pt]
\item { Step 1 \text{(feature detection and feature matching)}. We obtain matched features across pairs of images using a modern deep learning based feature detection and matching algorithm, GlueStick \cite{Pautrat2023gluestick}. Though we do not implement this in our experiments, there have been methods developed to further screen corrupted keypoint matches or obtain matches robustly, such as  \cite{shi2021scalable, shi2020robust, li2022fast}. Key points across a triplet of cameras is matched from pairs and is included only if it appears in all the pair combinations of the three images. }

\item Step 2 \text{(estimation and refinement of trifocal tensors)}. 
With the triplet matches, we calculate the trifocal tensors with more than 11 correspondences. To have an even sparser graph, one can skip the estimation of trifocal tensors and rely on the imputation for images that have less than a number bigger than 11 point correspondences. This can further speed up the trifocal tensor estimation process. 
We apply STE from \cite{yu2024subspace} to find 40\% of the correspondences as inliers, then use at most $30$ inlier point correspondences  to linearly estimate the trifocal tensor. 
To refine the estimates, we apply bundle adjustment on the inliers and delete triplets with reprojection error larger than 1 pixel. 


\item Step 3 \text{(synchronization)}. We synchronize the estimated block trifocal tensor with a robust variant of SVD using the framework described in Algorithm \ref{alg:sync}. The robustness comes from replacing SVD with a robust subspace recovery method \cite{yu2024subspace}. More details can be found in 
 \Cref{horste}. Recall that the cameras we retrieve are up to a global projective ambiguity. When comparing with ground truth poses, we first align our estimated cameras with the ground truth cameras by finding a $4\times 4$ projective transformation. Then we round the cameras to calibrated cameras and compare.
\end{itemize}

\begin{figure}[htp!]
    \centering    \includegraphics[width=\linewidth]{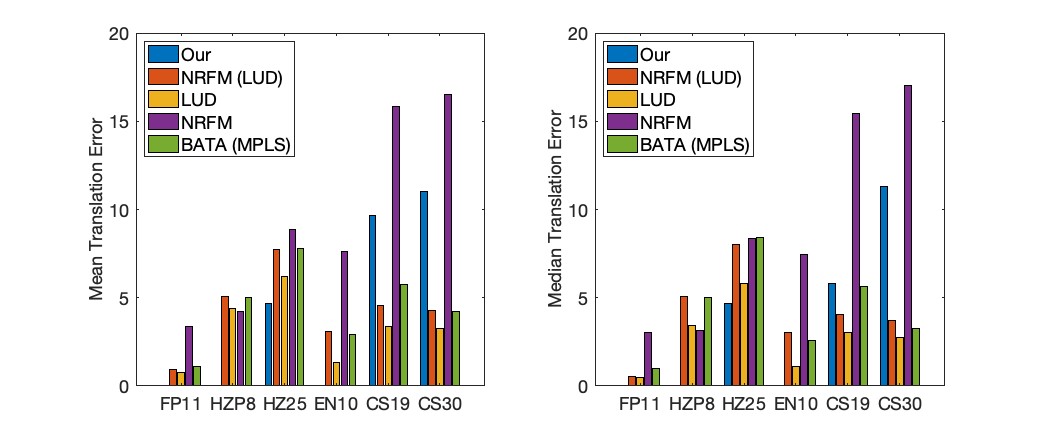}
    \caption{{EPFL translation error comparison between our method, NRFM initialized by LUD, LUD, and NRFM initialized randomly. BATA(MPLS) stands for BATA initialized by MPLS. HZ8 stands for HerzP8, FP11 for FountainP11, HZ25 for Herz P25, EN10 for EntryP10, CS19 for CastleP19, CS30 for CastleP30.}}
    \label{fig:epfl_results}
\end{figure}

We test our full pipeline on two EPFL datasets on a personal machine with 2 GHz Intel Core i5 with 4 cores and 16GB of memory. To test NRFM \cite{sengupta2017new}, LUD \cite{ozyesil2015robust} { and BATA \cite{Zhuang_2018_CVPR} initialized with MPLS \cite{shi2020message}}, we estimate the corresponding essential matrices using the { GC-ransac \cite{barath2018graph}}. We did not include blocks corresponding to two views in our trifocal tensor pipeline. The mean and median translation errors are summarized in Figure \ref{fig:epfl_results} here and { more comprehensive results can be found in }Table \ref{tab:EPFL_comprehensive} { and Table \ref{tab:EPFL_rotation_comprehensive}} in the appendix. 

{The EPFL datasets generally have a plethora of point correspondences, so that the trifocal tensors are estimated accurately. When the dataset focuses on a single scene, our algorithm retrieves locations competitively. Our algorithm achieves the best location estimation for 4 out of 6 datasets. The translation error bars are not visible for FP11, HZP8, EN10 due to the accuracy that we achieve. However, our pipeline is incapable of accurately processing CastleP19 and CastleP30. The main reason is that our algorithm relies on having a very dense observation graph to ensure high completion rate. CastleP19 and CastleP30 are datasets where the camera scans portions of the general area sequentially, so that not many triplets have overlapping features. Our method is not suitable for this type of dataset. However, it is possible to apply our algorithm in parallel on groups of neighboring frames, so that the completion rate is high in each group. Then the results can be merged to obtain a larger reconstruction. 
Rotations for the two view methods are estimated via rejecting outliers from iteratively applying \cite{Basri2012rotation}. We also compare against \cite{Zhuang_2018_CVPR} for location estimation, where we initialize with a state-of-the-art global rotation estimation method \cite{shi2020message}. Our algorithm achieves superior rotation estimation for only 2 out of the 6 datasets. See Table \ref{tab:EPFL_comprehensive} { and \ref{tab:EPFL_rotation_comprehensive} in the appendix for comprehensive errors. }   }

\subsection{Photo Tourism}
\begin{figure}[htp!]
    \centering    
    \begin{subfigure}{0.49\textwidth}
    \centering
    \includegraphics[width=\textwidth]{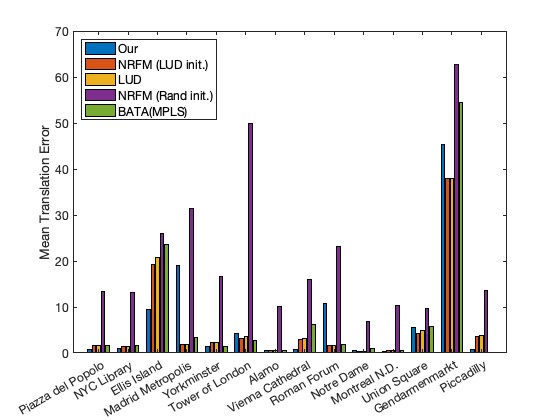}
    \caption{Photo Tourism mean translation errors}
    \label{fig:im22}
    \end{subfigure}    
    \hfill
    \begin{subfigure}{0.49\textwidth}
    \centering
    \includegraphics[width=\textwidth]{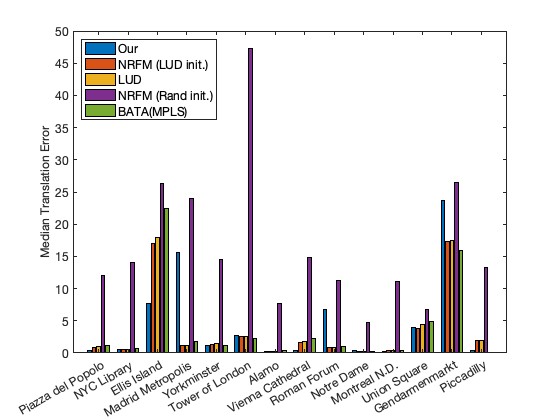}
    \caption{Photo Tourism median translation errors}
    \label{fig:im33}
    \end{subfigure}
    \caption{Photo Tourism translation error comparison between our method, NRFM initialized by LUD, LUD, NRFM initialized randomly, { and BATA initialized with MPLS. Note that we have not been able to acquire results for Piccadilly for BATA + MPLS}. }
    \label{fig:photo_results}
\end{figure}

We conduct experiments on the Photo Tourism datasets. The Photo Tourism datasets consist of internet images of real world scenes. Each scene has hundreds to thousands of images. The datasets \cite{1dsfm14} provide essential matrix estimates, and we estimate the trifocal tensors from the given essential matrices. 
To limit the computational cost for tensors, we downsample the datasets by choosing cameras with observations more than a certain percentage in the corresponding block frontal slice while maintaining a decent number of cameras. Note that this may not be the optimal way of extracting a dense subset in general. 
The maximum number of cameras we select for each dataset is 225 cameras. The largest dataset Piccadilly has 2031 cameras initially. We randomly sample 1000 cameras and then run our procedure. For Roman Forum and Piccadilly, the two view methods further deleted cameras from the robust rotation estimation process or parallel rigidity test. We rerun and report the trifocal tensor synchronization algorithm with the further downsampled data. We initialize the hard thresholding parameters for HOSVD-HT by first imputing the trifocal tensor with small random entries and then calculating the singular values for each of the flattenings. We take $l_i$ to be the tertile singular value for each mode-$i$ flattening. We then keep this parameter fixed for the synchronization process. Recall that the $jii$ blocks in the block trifocal tensor correspond to elements in the essential matrix $E_{ij}$. We also include these essential matrix estimations in the block trifocal tensor. The Photo Tourism experiments were run on an HPC center with 32 cores, { but the only procedure that can benefit from parallel computing in a single experiment is the scale retrieval}. Mean and median translation errors are summarized in Figure \ref{fig:photo_results}. Fully comprehensive results can be found in Tables \ref{tab:Photo_Comprehensive}  and \ref{tab:Photo_Comprehensive_rotations} in \Cref{comprehensive_results}. 

Our method is able to achieve competitive translation errors on 8 of the 14 datasets tested. 
{ Similar to the observation in the EPFL experiments, o}ur algorithm performs well when the viewing graph is dense, or in other words, when the estimation percentage is high. We achieve better locations in 6 out of 8 datasets where the estimation percentage exceeds 60\%, and better locations in only 2 out of 6 datasets where the estimation percentage falls below 60\%. We achieve reasonable rotation estimations for 10 out of 14 datasets, but not as good as LUD. See Table \ref{tab:Photo_Comprehensive_rotations} for a comprehensive result. Since the block trifocal tensor scales cubically with respect to the number of cameras, our algorithm runtime is longer than most two view global methods. This could be { alleviated} by synchronizing dense subsets in parallel and merging the results to construct a larger reconstruction.

\textbf{Additional remark:} Trifocal tensors can be estimated from line correspondences or a mix of point and line correspondences, while fundamental matrices are estimated from only point correspondences. There are many situations where accurate point correspondences are in short supply but there is a plethora of clear and distinct lines. { For example, see datasets in a recent SfM method using lines \cite{liu20233d}}. We demonstrate the potential of our method to be adapted to process datasets with only lines or very few points. Due to the limited availability of well annotated line datasets, we provide a small synthetic experiment that simulates a case where only lines correspondences are present. We first generate 20 random camera matrices, then we generate 25 lines that are projected on and shared across all images. We add about 0.02 percent of noise in terms of the relative { frobenius} norms between the line equation parameters and the noise. We estimate the trifocal tensor of three different views from line correspondences linearly. One remark is that our synchronization method works well only when the signs of the initial unknown scales are mostly uniform. We manually use ground truth trifocal tensors to correct the sign of the scale. { This has not been an issue in the previous experiments due to bundle adjustment for EPFL and the overall good estimations in Photo Tourism}. In practice, the sign of the scale on a trifocal tensor can be corrected via triangulation of points or reconstruction of lines, and correcting the sign using the depths of the reconstructed points or intersecting line segments. We synchronize the trifocal tensors with Algorithm \ref{alg:sync} and were able to achieve a mean rotation error of 0.61 degrees, median rotation error of 0.49 degrees, mean location error of 0.76, and median location error of 0.74. 






\section{Conclusion}\label{conclusion}
In this work, we introduced the block tensor of trifocal tensors characterizing the three-view geometry of a scene.  We established an explicit Tucker factorization of the block trifocal tensor and proved it has a low multilinear rank of $(6,4,4)$ under appropriate scaling.  We developed a synchronization algorithm based on tensor decomposition that retrieves an appropriate set of scales, and synchronizes rotations and translations simultaneously.  On several real data benchmarks we demonstrated state-of-the-art performance in terms of camera location estimation, and saw particular advantages on smaller and denser sets of images. Overall, this work suggests that higher-order interactions in synchronization problems have the potential to improve performance over pairwise-based methods. 

There are several limitations to our tensor-based synchronization method. First, our rotation estimations are not as strong as our location estimations. Second, our algorithm performance is affected by the estimation percentage of trifocal tensors within the block trifocal tensor. One could incorporate more robust completion methods and explore new approaches for processing sparse triplet graphs. Further, our block trifocal tensor scales cubically in terms of the number of cameras and becomes computationally expensive for large datasets. We can develop methods for extracting dense subgraphs, synchronizing in parallel, then merging results to obtain a larger reconstruction{, similarly to the distributed algorithms of \cite{Li2024efficient} and \cite{Dal2021synchronization}}. Moreover, our synchronization method's success depends on accurate trifocal tensor estimations, and it motivates further work on robust estimation of multi-view tensors. \Cref{alg:sync} could also be made more robust by adding outlier rejection techniques. Finally we plan to extend our theory by proving convergence of our algorithm and 
exploring structures for even higher-order tensors, such as quadrifocal tensors. 




\section*{Acknowledgement}
D.M.~and G.L.~were supported in part by NSF award DMS 2152766. J.K.~was supported in part by NSF awards DMS 2309782 and CISE-IIS 2312746, the DOE award SC0025312, and start-up grants from the College of Natural Science and Oden Institute at the University of Texas at Austin.

We thank Shaohan Li and Feng Yu for helpful discussions on processing EPFL and Photo Tourism. We also thank Hongyi Fan for helpful advice and references on estimating trifocal tensors.

\medskip

\bibliography{references}


\newpage
\appendix

\section{Appendix / supplemental material}
\subsection{Derivation of the trifocal tensor}\label{derivation_tft}
To provide a better intuition for the trifocal tensor, we briefly summarize the derivation of the trifocal tensor from
\citep{hartley2003multiple} and \citep{leonardos2015metric} under the general setup of uncalibrated cameras. 

Let $P_i = K_iR_i[I, -t_i]$ be the form of the camera matrix for $P_1,P_2,P_3$. Let $L$ be a line in the 3D world scene, and $l_1,l_2,l_3$ the corresponding projections in the images $I_1,I_2,I_3$ respectively. Each $l_i$ back projects to a plane $\pi_i = P_i^T l_i$ in $\mathbb{R}^3$, and since $l_i$ correspond to the same $L$ in the 3D world scene, $\pi = [\pi_1, \pi_2, \pi_3]$ must be rank deficient and its kernel will generically be spanned by $L$. 
Then, 
\begin{align*}
    \pi' = \begin{bmatrix}
    K_1^{-T}R_1 & 0\\
    t_1^{T} & 1
\end{bmatrix} \pi = \begin{bmatrix}
    l_1 & K_1^{-T}R_{12}K_2^Tl_2 & K_1^{-T}R_{13}K_3^Tl_3 \\
    0 & (t_1-t_2)^TR_2^TK_2^Tl_2 & (t_1-t_3)^TR_3^TK_3^Tl_3
\end{bmatrix} = [\pi'_1, \pi'_2, \pi'_3]
\end{align*}
will also be rank-deficient, implying that the columns of $\pi'$ are linearly dependent. This means that there exist $\alpha, \beta$ such that $\pi'_1 = \alpha \pi'_2 + \beta \pi'_3$. We can choose $\alpha = -(t_1-t_3)^TR_3^TK_3^Tl_3$, $\beta = (t_1-t_2)^TR_2^TK_2^Tl_2$, so that 
\begin{align*}
    l_1 = l_2^T [K_2 R_2 (t_1-t_2) K_1^{-T}R_{13}K_3^T]l_3 - l_2^T[K_2R_{12}^TK_1^{-1} (t_1-t_3)^TR_3^TK_3^T]l_3.
\end{align*}
Then, the \text{canonical trifocal tensor centered at camera 1} is defined as 
\begin{equation}\label{eq1}
     T_{i} = K_2 R_2 (t_1-t_2) e_i^T K_1^{-T}R_{13}K_3^T - K_2R_{12}^TK_1^{-1} e_i (t_1-t_3)^TR_3^TK_3^T
\end{equation}
where $e_i\in \R^3$ is the $i$-th standard basis vector. The trifocal tensor will be the tensor $\{T_1,T_2,T_3\}$, where the $T_i$'s are stacked along the first mode. The line incidence relation is then $(l_1)_i = l_2^T T_{i} l_3$.  Other combinations of point and line incidence relations are also encoded by the trifocal tensor; see \cite{hartley2003multiple} for details.  The construction for calibrated cameras is the same, just with $P_i$ in calibrated form.

\subsection{Proof details for  \Cref{theorem1}}\label{thrm1proof}
We include a detailed calculation for the Tucker factorization of the block trifocal tensor. Recall that each individual trifocal tensor corresponding to the cameras $a,b,c$ can be calculated as 
\begin{align*} 
   T_{iqr} =& (-1)^{i+1} \det\begin{bmatrix}
        \sim a^i\\
        b^q\\
        c^r
    \end{bmatrix} = (-1)^{i+1} \det\begin{bmatrix}
        a^m\\
        a^n\\
        b^q\\
        c^r
    \end{bmatrix} \\
    = & (-1)^{i+1}( 
    \det\begin{bmatrix}
        a_{m3} & a_{m4}\\
        a_{n3} & a_{n4}
    \end{bmatrix} (b_{q1}c_{r2} - b_{q2}c_{r1}) 
    +  \det\begin{bmatrix}
        a_{m2} & a_{m4}\\
        a_{n2} & a_{n4}
    \end{bmatrix} (-b_{q1}c_{r3} + b_{q3}c_{r1}) \\
    & \quad \quad + \det\begin{bmatrix}
        a_{m1} & a_{m4}\\
        a_{n1} & a_{n4}
    \end{bmatrix} (b_{q2}c_{r3} - b_{q3}c_{r2}) 
    +  \det\begin{bmatrix}
        a_{m2} & a_{m3}\\
        a_{n2} & a_{n3}
    \end{bmatrix} (b_{q1}c_{r4} - b_{q4}c_{r1}) \\  
    & \quad \quad + \det\begin{bmatrix}
        a_{m1} & a_{m3}\\
        a_{n1} & a_{n3}
    \end{bmatrix} (-b_{q2}c_{r4} + b_{q4}c_{r2}) + \det\begin{bmatrix}
        a_{m1} & a_{m2}\\
        a_{n1} & a_{n2}
    \end{bmatrix} (b_{q3}c_{r4} - b_{q4}c_{r3})
    ) \\
    =&  
    \p(a)_{i6} (b_{q1}c_{r2} - b_{q2}c_{r1})  
    + \p(a)_{i5} (-b_{q1}c_{r3} + b_{q3}c_{r1}) +  \p(a)_{i3} (b_{q2}c_{r3} - b_{q3}c_{r2}) \\
    & \quad +\p(a)_{i4} (b_{q1}c_{r4} - b_{q4}c_{r1}) +  \p(a)_{i2} (-b_{q2}c_{r4} + b_{q4}c_{r2}) +  \p(a)_{i1} (b_{q3}c_{r4} - b_{q4}c_{r3})\\
    =& \sum_{k=1}^6 \p(a)_{ik} \sum_{w=1}^4 b_{qw} \sum_{j=1}^4 c_{rj} \G_{kwj}.
\end{align*}

The last equality can be easily checked since $\G$ is sparse. For example, when $k=1$, $\p(a)_{i1}$ is the determinant of the submatrix dropping the $i$-th row and keeping columns 1 and 2, which is $\det \! \begin{bmatrix}
    a_{m1} a_{m2} ; a_{n1} a_{n2}
\end{bmatrix}$.
The only nonzero elements in the first horizontal slice are $\G(1,4,3) = -1$ and $\G(1,3,4) = 1$. Then, the nonzero elements in the sum when $k = 1$ will be exactly $\p(a)_{i1} \sum_{w=1}^4 b_{qw} \sum_{j=1}^4 c_{rj} \G_{1wj} = \p(a)_{i1} (b_{q3}c_{r4} - b_{q4}c_{r3})$.

Then, since $\p$ will be the stackings of $\p(P_i)$, $\C$ is the stacking of camera matrices in \Cref{theorem1}, each $ijk$ block in $T^n$ will be calculated by exactly the corresponding $i$, $j$, $k$ blocks in $\p, \C, \C$ respectively using the calculations above.

\subsection{Proof details for \Cref{prop1}}\label{prop1proof}
\begin{proof}[Proof for (i)]  We have
    \begin{align*}
    (T^n_{iii})_{wqr} = (-1)^{w+1} \det\begin{bmatrix}
    \sim P_i^w \\
    P_i^q \\
    P_i^r
\end{bmatrix} = 0,
\end{align*} 
since $P_i$ is a $3\times 4$ matrix and the submatrix above will always have two identical rows. 

\textit{Proof for (ii):}
Consider the $wqr$ element of the $jii$ block trifocal tensor, $T^n_{jii}$. It can be written as
    \begin{align*}
    (T^n_{jii})_{wqr} = (-1)^{w+1} \det\begin{bmatrix}
    \sim P_j^w \\
    P_i^q \\
    P_i^r
\end{bmatrix}.
\end{align*} 
Thus, when $q=r$, clearly $(T^n_{jii})_{wqr} = 0$ as we will have identical rows again. When $q\not = r$, we first observe that $(T^n_{jii})_{wqr} = -(T^n_{jii})_{wrq}$ since we just swap two rows. Second, 
    \begin{align*}
    (T^n_{jii})_{wqr} = (-1)^{w+1} \det\begin{bmatrix}
    \sim P_j^w \\
    P_i^q \\
    P_i^r
\end{bmatrix} = (-1)^{w+1} \det\begin{bmatrix}
    \sim P_j^w \\
    \sim P_i^m
\end{bmatrix} 
\end{align*} 
where $m \in \{1,2,3\} \setminus \{q,r\}$. This is exactly the bilinear relationship in \cite{hartley2003multiple} defining the fundamental matrix $(F_{ji})_{mw}$ element up to a possible negative sign.  

{\textit{Proof for (iii):} 
We can only show this for $T^n_{iij}$ blocks from symmetry. 
The elements in $T^n_{iij}$ blocks can be calculated as 
\begin{align*}
    (T^n_{iji})_{wqr} = (-1)^{w+1} \det \begin{bmatrix}
        \sim P_{i}^w \\ P_{i}^q \\ P_j^r
    \end{bmatrix}
\end{align*}

Elements are nonzero only when $w = q$, and they correspond to determinants of matrices with three rows from one $P_i$ and one row from $P_j$. By \cite{hartley2003multiple}, these are exactly the elements of the epipoles. When $w=1$, the order of the rows in the determinant corresponding to camera $i$ is $(2,3,1)$, when $w=2$, the order is $(1,3,2)$ and there is a negative sign in front of the determinant, and when $w=3$, the order is $(1,2,3)$. Since the first and last case are even permutations of the rows of $P_i$, and the second case is corrected by a negative sign, $(T^n_{iji})_{ww:}$ is exactly the epipole.  }

\textit{Proof for (iv):}   
On a horizontal slice, the camera along the 1st mode is fixed, and blocks symmetric across the diagonal is calculated by cameras, which the 2nd and 3rd mode cameras are swapped. Then, we will simply be swapping rows in \eqref{det def for tft}, which means that we will simply be changing signs for elements symmetric across the diagonal, implying skew symmetry. 
    
\textit{Proof for (v):}
Now assume that we have a block trifocal tensor whose corresponding cameras are all calibrated. Let $\p$ be the line projection matrix, $\C = [P_1, P_2, \dots, P_n]^T$ is the stacked camera matrix, and $\G$ is the core tensor. {The flattening in the 1st mode can be written as 
$T^n_{(1)} = \p \G_{(1)}(\C\otimes \C)^T$, where $T^n_{(1)}$ is a $3n \times 9n^2$ matrix.} For the proof, we calculate the eigenvalue of 
$T^n_{(1)} (T_{(1)}^n)^T = \p \G_{(1)}(\C\otimes \C)^T (\C\otimes \C) \G_{(1)}^T \p^T$
\begin{align*}
    T^n_{(1)} (T_{(1)}^n)^T &= \p \G_{(1)}(\C\otimes \C)^T (\C\otimes \C) \G_{(1)}^T \p^T \\
    &= \p \G_{(1)}(\C ^T \otimes \C^T) (\C\otimes \C) \G_{(1)}^T \p^T \\
    &= \p \G_{(1)}(\C ^T\C \otimes \C^T\C) \G_{(1)}^T \p^T.
\end{align*}
The second and third line uses two Kronecker product properties: $(A\otimes B)^T = A^T \otimes B^T$ and $(A \otimes B) (C \otimes D) = AC \otimes BD$ as long as $AC$ and $BD$ are defined. 

We first calculate $(\C^T\C \otimes \C^T\C)$. 

{ We assume that the cameras are centered at the origin, i.e. $\sum_{i=1}^n t_i = 0$. } Then we have 
\begin{align}
    \C^T\C = \begin{bmatrix}
    nI_{3\times 3}  &  -\sum_{i=1}^n t_i \\
    -\sum_{i=1}^n t_i^T & \sum_{i=1}^n \|t_i\|^2 
\end{bmatrix} = \begin{bmatrix}
    nI_{3\times 3}  &  0_{3\times 1} \\
    0_{1\times 3} & \sum_{i=1}^n \|t_i\|^2 
\end{bmatrix},
\end{align}
so that 
\begin{align}
    (\C^T\C \otimes \C^T\C) =& \begin{bmatrix}
    nI_{3\times 3}\otimes \C^T\C  &  0_{3\times 1}\otimes \C^T\C \\
    0_{1\times 3}\otimes \C^T\C & \sum_{i=1}^n \|t_i\|^2 \otimes \C^T\C 
\end{bmatrix} 
\end{align} 

We have an explicit form for $\G_{(1)}$:
\begin{align}
    \G_{(1)} = \begin{bmatrix} \begin{tabular}{cccccccccccccccc}
    0 & 0 & 0 & 0 & 0 & 0 & 0 & 0 & 0 & 0 & 0 & -1 & 0 & 0 & 1 & 0\\
    0 & 0 & 0 & 0 & 0 & 0 & 0 & 1 & 0 & 0 & 0 & 0 & 0 & -1 & 0 & 0\\
    0 & 0 & 0 & 0 & 0 & 0 & -1 & 0 & 0 & 1 & 0 & 0 & 0 & 0 & 0 & 0\\       
    0 & 0 & 0 & -1 & 0 & 0 & 0 & 0 & 0 & 0 & 0 & 0 & 1 & 0 & 0 & 0\\       
    0 & 0 & 1 & 0 & 0 & 0 & 0 & 0 & -1 & 0 & 0 & 0 & 0 & 0 & 0 & 0\\       
    0 & -1 & 0 & 0 & 1 & 0 & 0 & 0 & 0 & 0 & 0 & 0 & 0 & 0 & 0 & 0            
\end{tabular} \end{bmatrix}. 
\end{align}


Let $X = \C^T\C$ and let $X_{ij}$ denote the $ij$ entry in $\C^T\C$. Let $a = \sum_{i=1}^n \|t_i\|^2 $. 

We first show that $\G_{(1)}(\C^T\C \otimes \C^T\C)\G_{(1)}^T$ is diagonal by direct computation:
{\footnotesize
\begin{align*}
    &\G_{(1)}(\C^T\C \otimes \C^T\C) \G_{(1)}^T\\
=&\begin{bmatrix}
    nX_{44} + aX_{33} & aX_{22} & -nX_{42}& aX_{31}& nX_{41} & 0\\
    -aX_{23} & nX_{44} + aX_{22} & nX_{43} & aX_{21} & 0 & nX_{41}\\
    nX_{24} & -nX_{34} & nX_{33} + nX_{22} & 0 & -nX_{21} & -nX_{31}\\
    aX_{13} & -aX_{12} & 0 & nX_{44} + aX_{11} & -nX_{43} & nX_{42}\\
    nX_{14} & 0 & -nX_{12} & nX_{34} & nX_{33} + nX_{11} & -nX_{32}\\
    0 & -nX_{24} & -nX_{13} & nX_{24} & -nX_{23} & nX_{22}+nX_{11}
\end{bmatrix} \\
=&\begin{bmatrix}
    na + an & 0 & 0& 0& 0 & 0\\
    0 & na + an & 0 & 0 & 0 & 0\\
    0 & 0 & n^2 + n^2 & 0 & 0 & 0\\
    0 & 0 & 0 & na + an & 0 & 0\\
    0 & 0 & 0 & 0 & n^2 + n^2 & 0\\
    0 & 0 & 0 & 0 & 0 & n^2+n^2
\end{bmatrix} \\
=& \begin{bmatrix}
    2na & 0 & 0& 0& 0 & 0\\
    0 & 2na & 0 & 0 & 0 & 0\\
    0 & 0 & 2n^2 & 0 & 0 & 0\\
    0 & 0 & 0 & 2na & 0 & 0\\
    0 & 0 & 0 & 0 & 2n^2 & 0\\
    0 & 0 & 0 & 0 & 0 & 2n^2
\end{bmatrix}. 
\end{align*}}

We then calculate the spectral decomposition of $T^n_{(1)}$.
With a slight abuse of notation, let $P_i^k$ denote the $k$-th column of $P_i$. The $3n\times 6$ rank-$6$ stacked line projection matrix would have columns ordered according to 
\begin{align*}
    \begin{bmatrix}
    e_1 \wedge e_2 & e_1 \wedge e_3 &     e_1 \wedge e_4 &     e_2 \wedge e_3 &     e_2 \wedge e_4 &     e_3 \wedge e_4  
\end{bmatrix},
\end{align*} 

and since the second row in $\p$ for each camera is $P_i^3 \wedge P_i^1$ it holds

\begin{align*}
    \p = \begin{bmatrix}
    \vdots&\vdots&\vdots&\vdots&\vdots&\vdots \\
    P^1_i \times P^2_i & P^1_i \times P^3_i & P^1_i\times P^4_i & P^2_i \times P^3_i & P^2_i \times P^4_i &  P^3_i \times P^4_i \\   \vdots&\vdots&\vdots&\vdots&\vdots&\vdots
\end{bmatrix}.
\end{align*}

Or equivalently, the stacked wedge products between columns. Let $\p = USV^T$ be the thin singular value decomposition of $\p$, so that $U$ is a $3n\times 6$ orthonormal matrix, $S$ is a $6\times 6$ diagonal matrix where all diagonal entries are nonzero, and $V$ is a $6\times 6$ orthonormal matrix.

Then,
\begin{align*}
    T^n_{(1)} (T_{(1)}^n)^T &= \p \G_{(1)}(\C ^T\C \otimes \C^T\C) \G_{(1)}^T \p^T\\
              &= U(SV^T\G_{(1)}(\C ^T\C \otimes \C^T\C) \G_{(1)}^T V S) U^T.
\end{align*}

Since $V$ is orthonormal, $S V^T\G_{(1)}(\C ^T\C \otimes \C^T\C) \G_{(1)}^T V S$ is still a diagonal matrix. We just need to establish the fact that three of the diagonal entries are the same.

For one camera, $\p^T\p $ equals
{\scriptsize
\begin{align*}
& \begin{bmatrix}
        1 & 0 & P_i^2 \cdot P_i^4 & 0 & -P_i^1 \cdot P_i^4 & 0\\
          & 1 & P_i^3 \cdot P_i^4 & 0 & 0                  & -(P_i^1 \cdot P_i^4) \\
          &   & P_i^4 \cdot P_i^4- (P_i^1 \cdot P_i^4)(P_i^1 \cdot P_i^4) & 0 & - (P_i^1 \cdot P_i^4)(P_i^4 \cdot P_i^2) & - (P_i^1 \cdot P_i^4)(P_i^4 \cdot P_i^3)\\
          &   &                   & 1 & P_i^3 \cdot P_i^4  & - P_i^2 \cdot P_i^4\\
          &   &                   &   & (P_i^4 \cdot P_i^4) - (P_i^2 \cdot P_i^4)(P_i^4 \cdot P_i^2) & - (P_i^2 \cdot P_i^4)(P_i^4 \cdot P_i^3)\\
          &   &                   &   &                    & (P_i^4 \cdot P_i^4) - (P_i^3 \cdot P_i^4)(P_i^4 \cdot P_i^3)  
          \end{bmatrix} \\
=& \begin{bmatrix}
        1 & 0 & P_i^2 \cdot P_i^4 & 0 & -P_i^1 \cdot P_i^4 & 0\\
          & 1 & P_i^3 \cdot P_i^4 & 0 & 0                  & -(P_i^1 \cdot P_i^4) \\
          &   & \|P_i^4\|^2 -\|P_i^1\cdot P_i^4\|^2 & 0 & - (P_i^1 \cdot P_i^4)(P_i^4 \cdot P_i^2) & - (P_i^1 \cdot P_i^4)(P_i^4 \cdot P_i^3)\\
          &   &                   & 1 & P_i^3 \cdot P_i^4  & - P_i^2 \cdot P_i^4\\
          &   &                   &   & \|P_i^4\|^2 -\|P_i^2\cdot P_i^4\|^2 & - (P_i^2 \cdot P_i^4)(P_i^4 \cdot P_i^3)\\
          &   &                   &   &                    & \|P_i^4\|^2 -\|P_i^3\cdot P_i^4\|^2  
          \end{bmatrix}. 
\end{align*}
}
{ where the matrix is symmetric and we reduce redundancy by omitting the entries below the diagonal.}
For $n$ cameras, 
{\footnotesize
\begin{align*}
\p^T\p = \begin{bmatrix}
        n & 0 & \sum_i P_i^2 \cdot P_i^4 & 0 & - \sum_i P_i^1 \cdot P_i^4 & 0\\
          & n & \sum_i P_i^3 \cdot P_i^4 & 0 & 0                  & - \sum_i P_i^1 \cdot P_i^4  \\
          &   & \sum_i \|P_i^4\|^2 -\|P_i^1\cdot P_i^4\|^2 & 0 & - \sum_i (P_i^1 \cdot P_i^4)(P_i^4 \cdot P_i^2) & - \sum_i (P_i^1 \cdot P_i^4)(P_i^4 \cdot P_i^3)\\
          &   &                   & n & \sum_i P_i^3 \cdot P_i^4  & - \sum_i P_i^2 \cdot P_i^4\\
          &   &                   &   & \sum_i \|P_i^4\|^2 -\|P_i^2\cdot P_i^4\|^2 & - \sum_i (P_i^2 \cdot P_i^4)(P_i^4 \cdot P_i^3)\\
          &   &                   &   &                    & \sum_i \|P_i^4\|^2 -\|P_i^3\cdot P_i^4\|^2  
          \end{bmatrix}. 
\end{align*}
}

{ where $P_i^a \cdot P_i^b$ means the dot product between the $a$th and $b$th column}

\!\! The SVD of $\p^T\p$ is $\p^T\p = HDH^T$, where $H$ is $6\times 6$ orthonormal matrix, $D$ is $6\times 6$ diagonal matrix. { However, since we have an $nI_{3\times 3}$ submatrix in $\p^T\p$, we deduce that n appears as an eigenvalue 3 times for $\p^T \p$, where we can use the determinant identity for block matrices. We check that this indeed holds by a computer calculation, generating random instances of $P_i$'s and calculating the eigenvalues for $\p^T\p$. }


As a result, in the thin SVD of $\p$, we have $\p = USV^T$ where $S = \sqrt{D}$, $V = H$. Then in
\begin{align*}
    T^n_{(1)} (T^n_{(1)})^T &= \p \G_{(1)}(\C ^T\C \otimes \C^T\C) \G_{(1)}^T \p^T\\
              &= U(SV^T \G_{(1)}(\C ^T\C \otimes \C^T\C) \G_{(1)}^T V S) U^T,
\end{align*}

we see that $SV^T\G_{(1)}(\C^T\C \otimes \C^T\C) \G_{(1)}^T V S$ is a diagonal matrix where three of the entries are the same. By the uniqueness of the eigenvalues, we see that we have a spectral decomposition of $T^n_{(1)}(T^n_{(1)})^T$, so that three of the singular values of $T^n_{(1)}$ are equal. 
\end{proof}







\subsection{Proof details for \Cref{thm:2}}\label{app:proof2}
\begin{proof}
Note blockwise multiplication by a rank-$1$ tensor with nonzero entries preserves multilinear rank, since it is a Tucker product by invertible diagonal matrices. 
Therefore, without loss of generality we may assume $\lambda_{i11} = \lambda_{1j1} = \lambda_{11k} = 1$ for all $i,j,k \in \{2, 3, \ldots, n\}$.
Below we will prove it follows $\lambda_{ijk} = c$ if exactly one of $i,j,k$ equals $1$, and $\lambda_{ijk} = c^2$ if none of $i,j,k$ equal $1$ and the indices are not all the same, for some constant $c \in \mathbb{R}^*$.
This will immediately imply the theorem, because taking $\alpha = \beta = (1 \, c \, c \,  \ldots \, c)$ and $\gamma = (\tfrac{1}{c} \, 1 \, 1 \, \ldots \, 1)$ achieves $\lambda_{ijk} = \alpha_i \beta_j \gamma_k$ whenever $i,j,k$ are not all the same.

We consider the matrix flattenings $T^n_{(2)}$ and $(\lambda \odot_b T^n)_{(2)}$ in $\R^{3n \times 9n^2}$ of the block trifocal tensor and its scaled counterpart, 
with rows corresponding to the second mode of the tensors.  
By 
\Cref{theorem1} and assumptions, the flattenings have matrix rank $4$, thus all of their $5 \times 5$ minors vanish.
The argument consists of considering several carefully chosen $5 \times 5$ submatrices of $(\lambda \odot_b T^n)_{(2)}$ to prove the existence of a constant $c$ as above.
Index the rows and columns of the flattenings by $(j,r)$ and $(iq, ks)$ respectively, for $i,j,k \in [n]$ and $q,r,s \in [3]$, so that e.g.,  
 $((\lambda \odot_b T^n)_{(2)})_{(j,r), (iq, ks)} = \lambda_{ijk} (T^n_{ijk})_{qrs}$.

\textbf{Step 1:}  The first submatrix of $(\lambda \odot_b T^n)_{(2)}$ we consider has column labels 
$(i1,11)$, $(i1,21)$, $(i1,31)$, $(i1,12)$, $(i2,11)$ and 
row labels $(1,1)$, $(1,2), (1,3), (i,1), (i,2)$, where $i \in \{2, \ldots, n\}$.  Explicitly, it is 
\begin{equation*}
\begin{bmatrix}
(T^n_{i11})_{111} & (T^n_{i11})_{211} & (T^n_{i11})_{311} & (T^n_{i11})_{112} & (T^n_{11i})_{111} \\[1.5pt] 
(T^n_{i11})_{121} & (T^n_{i11})_{221} & (T^n_{i11})_{321} & (T^n_{i11})_{122} & (T^n_{11i})_{121} \\[1.5pt]
(T^n_{i11})_{131} & (T^n_{i11})_{231} & (T^n_{i11})_{331} & (T^n_{i11})_{132} & (T^n_{11i})_{131} \\[1.5pt]
\lambda_{ii1}(T^n_{ii1})_{111} & \lambda_{ii1}(T^n_{ii1})_{211} & \lambda_{ii1}(T^n_{ii1})_{311} & \lambda_{ii1}(T^n_{ii1})_{112} & \lambda_{1ii}(T^n_{1ii})_{111} \\[1.5pt]
\lambda_{ii1}(T^n_{ii1})_{121} & \lambda_{ii1}(T^n_{ii1})_{221} & \lambda_{ii1}(T^n_{ii1})_{321} & \lambda_{ii1}(T^n_{ii1})_{122} & \lambda_{1ii}(T^n_{1ii})_{121}
\end{bmatrix},
\end{equation*}
which we abbreviate as 
\begin{equation} \label{eq:det1}
\begin{bmatrix}
\ast & \ast & \ast & \ast & \ast \\
\ast & \ast & \ast & \ast & \ast \\
\ast & \ast & \ast & \ast & \ast \\
\lambda_{ii1}\ast & \lambda_{ii1}\ast & \lambda_{ii1}\ast & \lambda_{ii1}\ast & \lambda_{1ii}\ast \\
\lambda_{ii1}\ast & \lambda_{ii1}\ast & \lambda_{ii1}\ast & \lambda_{ii1}\ast & \lambda_{1ii}\ast \\
\end{bmatrix},
\end{equation}
with asterisk denoting the corresponding entry in $T^n_{(2)}$.  
As a function of $\lambda_{ii1}, \lambda_{1ii}$, the determinant of \eqref{eq:det1} is a degree $\leq 2$ polynomial, which must be divisible by $\lambda_{ii1}$ and $\lambda_{ii1} - \lambda_{1ii}$ (because if $\lambda_{ii1} = 0$ then clearly the bottom two rows of \eqref{eq:det1} are linearly independent, and if $\lambda_{ii1} - \lambda_{1ii} = 0$ we have a submatrix of $T^n_{(2)}$ with the bottom two rows scaled uniformly).  So the determinant of $\eqref{eq:det1}$ is a scalar multiple of $\lambda_{ii1}(\lambda_{ii1} - \lambda_{1ii})$. 
Note that the multiple is a polynomial function of the cameras $P_1$ and $P_i$.  
We claim that generically the multiple is nonzero; and to see this, it suffices to exhibit a \textit{single} instance of (calibrated) cameras where the determinant of \eqref{eq:det1} does not vanish identically for all $\lambda_{ii1}, \lambda_{1ii}$ due to the polynomiality (e.g., see \cite{harris1992algebraic}).  We check that this indeed holds by a computer calculation, generating numerical instances of $P_1$ and $P_i$ randomly.  
Thus the vanishing of the minor in \eqref{eq:det1} implies $\lambda_{ii1}(\lambda_{ii1} - \lambda_{1ii}) = 0$, whence $\lambda_{ii1} = \lambda_{1ii}$ since $\lambda_{ii1} \neq 0$.
An analogous calculation with $(\lambda \odot_b T^n)_{(3)}$ gives $\lambda_{i1i} = \lambda_{1ii}$.

\smallskip

\textbf{Step 2:} Next consider the submatrix of $(\lambda \odot_b T^n)_{(2)}$ with column labels $(j1,11)$, $(j1,21)$, $(j1,31)$, $(j1,12)$, $(1j,11)$ and row labels $(1,1)$, $(1,2)$, $(1,3)$, $(i,1)$, $(i,2)$, where $i,j \in \{2, \ldots, n\}$ are distinct.  It looks like  
\begin{equation} \label{eq:det2}
\begin{bmatrix}
\ast & \ast & \ast & \ast & \ast \\
\ast & \ast & \ast & \ast & \ast \\
\ast & \ast & \ast & \ast & \ast \\
\lambda_{ji1}\ast & \lambda_{ji1}\ast & \lambda_{ji1}\ast & \lambda_{ji1}\ast & \lambda_{1ij}\ast \\
\lambda_{ji1}\ast & \lambda_{ji1}\ast & \lambda_{ji1}\ast & \lambda_{ji1}\ast & \lambda_{1ij}\ast \\
\end{bmatrix},
\end{equation}
with asterisks denoting entries of $T^n_{(2)}$.
Similarly to the previous case, the determinant of \eqref{eq:det2} must be a scalar multiple of $\lambda_{ji1}(\lambda_{ji1} - \lambda_{1ij})$ where the scale depends polynomially on $P_1, P_i, P_j$.  By a computer computation, we find that the scale is nonzero for random instances of cameras (alternatively, note the polynomial system in step 1 is a special case of the present one).  It the scale is generically nonzero, hence $\lambda_{ji1} = \lambda_{1ij}$.
An analogous calculation with $(\lambda \odot_b T^n)_{(3)}$ gives $\lambda_{i1j} = \lambda_{1ij}$.

\smallskip

\textbf{Step 3:} Consider the submatrix of $(\lambda \odot_b T^n)_{(2)}$ with column labels $(j1,11)$, $(j1,21)$, $(j1,31)$, $(j1,12)$, $(1k,11)$ and row labels $(1,1)$, $(1,2)$, $(1,3)$, $(i,1)$, $(i,2)$, for $i,j,k \in \{2, \ldots, n\}$ distinct.  It looks like
\begin{equation} \label{eq:det3}
\begin{bmatrix}
\ast & \ast & \ast & \ast & \ast \\
\ast & \ast & \ast & \ast & \ast \\
\ast & \ast & \ast & \ast & \ast \\
\lambda_{ji1}\ast & \lambda_{ji1}\ast & \lambda_{ji1}\ast & \lambda_{ji1}\ast & \lambda_{1ik}\ast \\
\lambda_{ji1}\ast & \lambda_{ji1}\ast & \lambda_{ji1}\ast & \lambda_{ji1}\ast & \lambda_{1ik}\ast \\
\end{bmatrix}.
\end{equation}
The determinant of \eqref{eq:det3} is a scalar multiple of $\lambda_{ji1} (\lambda_{ji1} - \lambda_{1ik})$. 
By a direct computer computation as before, it is a nonzero multiple generically (alternatively, note the polynomial system in step 1 is a special of the present one). 
We deduce $\lambda_{ji1} = \lambda_{1ik}$.  
An analogous calculation with $(\lambda \odot_b T^n)_{(3)}$ gives $\lambda_{i1j} = \lambda_{1kj}$.

In particular, combining with step 2 it follows $\lambda_{1ij} = \lambda_{1ji}$, because $\lambda_{1ij} = \lambda_{k1j} = \lambda_{1kj} = \lambda_{ik1} = \lambda_{1ki} = \lambda_{j1i} = \lambda_{1ji}$.  From this, step 1 and step 2, we have that the $\lambda$-scale does not depend on the ordering of its indices, provided there is a $1$ among the indices.

\smallskip

\textbf{Step 4:} Consider the submatrix of $(\lambda \odot_b T^n)_{(2)}$ with column labels $(j1,11)$, $(j1,21)$, $(j1,31)$, $(j1,12)$, $(1i,11)$ and row labels
$(1,1)$, $(1,2)$, $(1,3)$, $(i,1)$, $(i,2)$, for $i, j \in \{2, \ldots, n\}$ distinct.  It looks like
\begin{equation} \label{eq:det4}
\begin{bmatrix}
\ast & \ast & \ast & \ast & \ast \\
\ast & \ast & \ast & \ast & \ast \\
\ast & \ast & \ast & \ast & \ast \\
\lambda_{ji1}\ast & \lambda_{ji1}\ast & \lambda_{ji1}\ast & \lambda_{ji1}\ast & \lambda_{1ii}\ast \\
\lambda_{ji1}\ast & \lambda_{ji1}\ast & \lambda_{ji1}\ast & \lambda_{ji1}\ast & \lambda_{1ii}\ast \\
\end{bmatrix}.
\end{equation}
The determinant of \eqref{eq:det4} is a scalar multiple of $\lambda_{ji1}(\lambda_{ji1} - \lambda_{1ii})$.  By a direct computer computation, it is a nonzero multiple generically (alternatively, note  the polynomial system in step 1 is a special case of the present one).  We deduce $\lambda_{ji1} = \lambda_{1ii}$.  

Putting together what we know so far, all $\lambda$-scales with a single $1$-index agree.  Indeed, this follows from $\lambda_{1ii} = \lambda_{ji1} = \lambda_{ij1} = \lambda_{1jj}$ so all $\lambda$-scales with a single $1$-index and two repeated indices agree, combined with $\lambda_{ji1} = \lambda_{1ii}$ and the last sentence of step 3.  Let $c \in \mathbb{R}^*$ denote this common scale.

\smallskip

\textbf{Step 5:} Consider the submatrix of $(\lambda \odot_b T^n)_{(2)}$ with column labels 
$(1i,11)$, $(1i,21)$, $(1i,31)$, $(1i,12)$, $(ij,11)$ and row labels 
$(1,1)$, $(1,2)$, $(1,3)$, $(i,1)$, $(i,2)$, for $i, j \in \{2, \ldots, n\}$ distinct.
It looks like 
\begin{equation} \label{eq:det5}
\begin{bmatrix}
\ast & \ast & \ast & \ast & c\ast \\
\ast & \ast & \ast & \ast & c\ast \\
\ast & \ast & \ast & \ast & c\ast \\
c\ast & c\ast & c\ast & c\ast & \lambda_{iij}\ast \\
c\ast & c\ast & c\ast & c\ast & \lambda_{iij}\ast \\
\end{bmatrix}.
\end{equation}
As a function of $c$ and $\lambda_{iij}$, the determinant of \eqref{eq:det5} is a scalar multiple of $c(c^2 - \lambda_{iij})$ (the second factor is present because it corresponds to scaling the bottom two rows and rightmost column of a $5 \times 5$ submatrix of $T_{(2)}$ each by $c$, which preserves rank deficiency).  By a direct computer computation, we find that the scale is nonzero for a random instance of $P_1, P_i, P_j$, therefore it is nonzero generically.  It follows $c^2 = \lambda_{iij}$.  An analogous calculation with $(\lambda \odot_b T^n)_{(3)}$ gives $c^2 = \lambda_{iji}$. 
\smallskip

\textbf{Step 6:} Consider the submatrix of $(\lambda \odot_b T^n)_{(2)}$ with column labels 
$(1i,11)$, $(1i,21)$, $(1i,31)$, $(1i,12)$, $(ji,11)$ and 
row labels $(1,1)$, $(1,2)$, $(1,3)$, $(i,1)$, $(i,2)$, for $i, j \in \{2, \ldots, n\}$ distinct.
It looks like 
\begin{equation} \label{eq:det6}
\begin{bmatrix}
\ast & \ast & \ast & \ast & c\ast \\
\ast & \ast & \ast & \ast & c\ast \\
\ast & \ast & \ast & \ast & c\ast \\
c\ast & c\ast & c\ast & c\ast & \lambda_{jii}\ast \\
c\ast & c\ast & c\ast & c\ast & \lambda_{jii}\ast \\
\end{bmatrix}.
\end{equation}
Similarly to the previous step, the determinant of \eqref{eq:det6} must be a scalar multiple of $c (c^2 - \lambda_{jii})$.  By a direct computer computation, it is a nonzero multiple generically.  We deduce $c^2 = \lambda_{jii}$.

\smallskip 

\textbf{Step 7:} Consider the submatrix of $(\lambda \odot_b T^n)_{(2)}$ with column labels 
$(1i,11)$, $(1i,21)$, $(1i,31)$, $(1i,12)$, $(ik,11)$ and row labels $(1,1)$, $(1,2)$, $(1,3)$, $(j,1)$, $(j,2)$, for $i, j, k \in \{2, \ldots, n\}$ distinct.
It looks like 
\begin{equation} \label{eq:det7}
\begin{bmatrix}
\ast & \ast & \ast & \ast & c\ast \\
\ast & \ast & \ast & \ast & c\ast \\
\ast & \ast & \ast & \ast & c\ast \\
c\ast & c\ast & c\ast & c\ast & \lambda_{ijk}\ast \\
c\ast & c\ast & c\ast & c\ast & \lambda_{ijk}\ast \\
\end{bmatrix}.
\end{equation}
The determinant of \eqref{eq:det7} is a scalar multiple of $c(c^2 - \lambda_{ijk})$.  By a direct computer computation, it is a nonzero multiple generically (alternatively, note the polynomial system in step 5 is a special case of the present case).  We deduce $c^2 = \lambda_{ijk}$.

At this point, by steps 5,6,7 we have that all $\lambda$-scales with no $1$-indices and  not all indices the same must equal $c^2$.  Combined with the second paragraph of step 4, this shows $c$ satisfies the property announced at the start of the proof.  Therefore the proof is complete.
\end{proof}
\subsection{Implementation details}

\subsubsection{Estimating trifocal tensors from three fundamental matrices}\label{tft_from_ess}

Given three cameras $P_1,P_2,P_3$ and the corresponding fundamental matrices $F_{21}, F_{31}, F_{32}$, we can calculate the trifocal tensor $T_{ijk}$ using the following procedure detailed in \cite{hartley2003multiple}. Specifically, from $F_{21}$ calculate an initial estimate of the cameras $P_1', P_2'$. Then, $P_3^T F_{32} P_2'$ and $P_3^T F_{31} P_1'$ should be skew-symmetric matrices. This gives 20 linear equations in terms of the entries in $P_3$, which can be used to solve for the trifocal tensor. Note that there are no geometrical constraints when calculating $P_3$, and there will be no guarantee of the quality of the estimation. 

\subsubsection{Higher-order regularized subspace-constrained Tyler's estimator (HOrSTE) for EPFL}\label{horste}

We describe the robust variant of SVD that we used for the EPFL experiments in  \Cref{num_sec}. Numerically, it performs more stably and accurately than HOSVD-HT, yet it is an iterative procedure and each iteration requires an SVD of the $3n \times 9n^2$ flattening. This becomes computationally expensive when $n$ becomes large and the number of iterations are also large. However, since the number of cameras for the EPFL dataset are below 20 cameras, the computational overhead is not too great. 

In HOSVD, a low dimensional subspace is estimated using the singular value decomposition and taking the $R_n$ leading left singular vectors for each mode-$n$ flattening. The Tyler's M Estimator (TME) \cite{tyler1987distribution} is a robust covariance estimator of a $D$ dimensional dataset $\{x_i\}_{i=1}^N \subset \R^D$. It minimizes the objective function 
\begin{equation}
\underset{\Sigma \in \R^{D\times D}}{\min} \frac{D}{N} \sum_{i=1}^N \log(x_i^T \Sigma^{-1} x_i) + \log \det(\Sigma)  
\end{equation}
such that $\Sigma$ is positive definite and has trace 1. The TME estimator can be applied to robustly find an $R_n$ dimensional subspace by taking the $R_n$ leading eigenvectors of the covariance matrix of TME. 
To compute the TME, \cite{tyler1987distribution} proposes an iterative algorithm, where 
\begin{equation}
    \Sigma^{(k)} \, = \, \sum_{i=1}^N \frac{x_ix_i^T}{x_i^T(\Sigma^{(k-1)})^{-1})x_i} / tr(\sum_{i=1}^N \frac{x_ix_i^T}{x_i^T(\Sigma^{(k-1)})^{-1})x_i}).
\end{equation}
TME doesn't exist when $D>N$, but a regularized TME has been proposed by \cite{sun2014regularized}. The iterations become 
\begin{equation}
    \Sigma^{(k)} = \frac{1}{1+\alpha} \frac{D}{N}\sum_{i=1}^N \frac{x_ix_i^T}{x_i^T(\Sigma^{(k-1)})^{-1})x_i}  + \frac{\alpha}{1+\alpha} I
\end{equation}
where $\alpha$ is a regularization parameter, and $I$ is the $D\times D$ identity matrix. 
TME does not assume the dimension of the subspace $d$ is predetermined. In the case when $d$ is prespecified, \cite{yu2024subspace} improves the TME estimator by incorporating the information into the algorithm and develops the subspace-constrained Tyler's Estimator (STE). For each iteration, STE equalizes the trailing $D-d$ eigenvalues of the estimated covariance matrix and uses a parameter $0<\gamma<1$ to shrink the eigenvalues. The iterative procedure for STE is summarized into 3 steps: 
\begin{enumerate}\label{STE_procedure}
    \item Calculate the unnormalized TME, $Z^{(k)} =  \sum_{i=1}^N (x_ix_i^T/x_i^T(\Sigma^{(k-1)})^{-1})x_i)$. 
    \item Perform the eigendecomposition of $Z^{(k)}= U^{(k)} S^{(k)} (U^{(k)})^T$, and set the trailing $D-d$ eigenvalues as $\gamma \sum_{i=d+1}^D \sigma_i/(D-d)$.
    \item Calculate $\Sigma^{(k)} = U^{(k)} S^{(k)} (U^{(k)})^T/tr(U^{(k)} S^{(k)} (U^{(k)})^T)$, which is the normalized covariance matrix. Repeat steps 1-3 until convergence. 
\end{enumerate}

Similar to the regularized TME, STE can also be regularized to succeed in situations where there are fewer inliers, and can improve the robustness of the algorithm. The \textit{regularized STE} differs from STE in only the first step, which is replaced by  \begin{enumerate}
    \item[1.*] Calculate the unnormalized regularized TME,   $Z^{(k)} =  \frac{1}{1+\alpha} \frac{D}{N}\sum_{i=1}^N \frac{x_ix_i^T}{x_i^T(\Sigma^{(k-1)})^{-1})x_i}  + \frac{\alpha}{1+\alpha} I$
\end{enumerate}

We apply the regularized STE to the HOSVD framework, and call the resulting projection the \textit{higher-order regularized STE} (HOrSTE). It is performed via the following steps:
\begin{enumerate}
    \item For each $n$, calculate the factor matrices $A_n$ as the $R_n$ leading left singular vectors from regularized STE applied to $T_{(n)}$. 
    \item \text{ Set the core tensor } $\G$ \text{ as } $\G = T \times_1 A_1^T \times_2 A_2^T \cdots \times_N A_N^T$.
\end{enumerate}


\subsection{Additional numerical results}\label{comprehensive_results}
In this section, we include comprehensive results for the rotation and translation errors for the EPFL and Photo Tourism experiments. Table \ref{tab:EPFL_comprehensive} { and \ref{tab:EPFL_rotation_comprehensive} contains all results for EPFL datasets}. Table \ref{tab:Photo_Comprehensive} contains the location estimation errors for Photo Tourism. Table \ref{tab:Photo_Comprehensive_rotations} contains the rotation estimation errors for Photo Tourism. In Table \ref{tab:Photo_Comprehensive_rotations}, we only report the rotation errors for LUD for all the methods that we compared against, as they are mostly the same since they used the same rotation averaging method.

\begin{table}[htp!]
  \caption{{EPFL synchronization errors. $\bar{e}_r$ is the mean rotation error in degrees, $\hat{e}_r$ is the median rotation error in degrees. $\bar{e}_t$ is the mean location error, $\hat{e}_t$ is the median location error. NRFM(LUD) is NRFM initialized with LUD and NRFM is randomly initialized. BATA(MLPS) is BATA initialized with MPLS. }}
    \label{tab:EPFL_comprehensive}
    \centering
    \tabcolsep=0.04cm
\begin{tabular}{||c | c  c |c c |c c |cc|cc||} 
 \hline
 &\multicolumn{2}{|c|}{Our} &\multicolumn{2}{|c|}{LUD} &\multicolumn{2}{|c|}{NRFM(LUD)} &\multicolumn{2}{|c|}{NRFM} &\multicolumn{2}{|c||}{BATA(MPLS)}\\
 \hline
   Dataset &  $\bar{e}_t$ & $\hat{e}_t$ & $\bar{e}_t$ & $\hat{e}_t$  & $\bar{e}_t$ & $\hat{e}_t$  & $\bar{e}_t$ & $\hat{e}_t$ & $\bar{e}_t$ & $\hat{e}_t$\\ [0.5ex] 
 \hline\hline
 FountainP11 & \textbf{0.008} & \textbf{0.007} & 0.91 & 0.54 & 0.75 & 0.46  & 3.37 & 3.03 & 1.12 & 1.01\\
\hline
HerzP8 &   \textbf{0.02} & \textbf{0.02}  & 5.06 & 5.06 & 4.37 & 3.42 & 4.24 & 3.14  & 5.04 & 5.03\\
 \hline
 HerzP25 & \textbf{4.70} & \textbf{4.68}  & 7.75 & 8.00 & 6.20 & 5.82 & 8.85 & 8.38  & 7.77 & 8.41\\
 \hline
  EntryP10  & \textbf{0.05} & \textbf{0.02}  & 3.08 & 3.02 & 1.34 & 1.11 & 7.63 & 7.43  & 2.90 & 2.58\\
 \hline
  CastleP19  & 9.64 & 5.80 & 4.58 & 4.04 & \textbf{3.37} & \textbf{3.02} & 15.81 & 15.43 & 5.77 & 5.62\\
 \hline
  CastleP30 & 11.00 & 11.33  & 4.27 & 3.72 & \textbf{3.24} & \textbf{2.75} & 16.54 & 17.04 & 4.23 & 3.26\\
 \hline \hline
\end{tabular}
\end{table}

\begin{table}[htp!]
  \caption{{EPFL synchronization errors. $\bar{e}_r$ is the mean rotation error in degrees, $\hat{e}_r$ is the median rotation error in degrees. BATA(MLPS) is BATA initialized with MPLS. }}
    \label{tab:EPFL_rotation_comprehensive}
    \centering
    \tabcolsep=0.04cm
\begin{tabular}{||c| c c  |c c |cc||} 
 \hline
 &\multicolumn{2}{|c|}{Our} &\multicolumn{2}{|c|}{LUD} &\multicolumn{2}{|c||}{BATA(MPLS)}\\
 \hline
   Dataset &  $\bar{e}_r$ & $\hat{e}_r$ &  $\bar{e}_r$ & $\hat{e}_r$  & $\bar{e}_r$ & $\hat{e}_r$ \\ [0.5ex] 
 \hline\hline
 FountainP11 & 0.09 & 0.08  & \textbf{0.05} & \textbf{0.05} & 0.06 & 0.05 \\
\hline
HerzP8 &  \textbf{0.12} & \textbf{0.12}  & 0.33 & 0.34 & 0.44 & 0.39  \\
 \hline
 HerzP25 & 2.01 & 1.11 & \textbf{0.18} & \textbf{0.19} & 0.26 & 0.23 \\
 \hline
  EntryP10 & \textbf{0.15} & \textbf{0.11} & 0.25 & 0.25 & 0.27 & 0.25 \\
 \hline
  CastleP19 & 56.24 & 11.71 & \textbf{0.24} & \textbf{0.22} & 0.27 & 0.25 \\
 \hline
  CastleP30 &  38.84 & 4.58 & \textbf{0.13} & \textbf{0.13} & 0.19 & 0.15 \\
 \hline \hline
\end{tabular}
\end{table}

\begin{table}[htp!]
    \caption{ { Translation errors for Photo Tourism. $n$ is the size after downsampling. Est. \% is the ratio of observed blocks over total number of blocks. $\bar{e}_t$ is the mean location error, $\hat{e}_t$ is the median location error. NRFM(L) is NRFM initialized with LUD and NRFM(R) is randomly initialized. The notation PR means that the dataset was further downsampled to match the two view methods. BATA is BATA initialized with MPLS. We were not able to get results for our subsampled dataset for Piccadilly with MPLS. }}
    \label{tab:Photo_Comprehensive}
    \centering
     \tabcolsep=0.055cm
\begin{tabular}{||c | c c c c |c c |c c |c c |cc||} \hline
Dataset &\multicolumn{4}{|c|}{Our Approach} &\multicolumn{2}{|c|}{NRFM(L)} &\multicolumn{2}{|c|}{LUD}&\multicolumn{2}{|c|}{NRFM(R)}&\multicolumn{2}{|c||}{BATA}\\
 \hline
   dataset & n & Est. \% &  $\bar{e}_t$ & $\hat{e}_t$ &  $\bar{e}_t$ & $\hat{e}_t$&  $\bar{e}_t$ & $\hat{e}_t$&  $\bar{e}_t$ & $\hat{e}_t$&  $\bar{e}_t$ & $\hat{e}_t$\\
 \hline\hline
  {Piazza del Popolo} & 185 & 72.3  & \textbf{0.78} &\textbf{ 0.45} & 1.63 & 0.85 & 1.66 & 0.86 & 13.45 & 12.06 & 1.63 & 1.10\\
  \hline
  {NYC Library} & 127 & 64.7 & \textbf{1.01}& \textbf{0.53} &  1.39 & 0.48 & 1.49 & 0.57 & 13.06 & 14.03 & 1.59 & 0.68\\
  \hline
 {Ellis Island } & 194 & 70.3  & \textbf{9.56} &  \textbf{7.73} & 19.31 & 16.97 & 20.71 & 17.96 & 26.08 & 26.38 & 23.63 & 22.50\\
 \hline
  {Tower of London} & 130 & 34.1 & 4.15 & 2.66 & 3.26 & 2.49 & 3.54 & 2.51 & 49.99 & 47.33 & \textbf{2.70} & \textbf{2.26}\\
 \hline
{Madrid Metropolis} & 190 & 35.9  & 18.93 & 15.53 & \textbf{1.91} &\textbf{ 1.19} & 1.94 & 1.20 & 31.48 & 24.02 & 3.33 & 1.72\\
  \hline
   {Yorkminster} &196 & 37.2 & 1.46& \textbf{1.14} & 2.31 & 1.39 & 2.35 & 1.45 & 16.67 & 14.46 & \textbf{1.37} & 1.15\\
  \hline
{Alamo}  & 224 & 94.3 & 0.62& \textbf{0.28} & \textbf{0.53} & 0.31 & \textbf{0.53} & 0.31 & 10.04 & 7.68 & 0.55 & 0.33\\
\hline
{Vienna Cathedral}& 197& 97.8  &\textbf{ 0.73} & \textbf{0.33} & 2.96 & 1.64 & 3.15 & 1.79 & 16.08 & 14.76 & 6.16 & 2.18\\
\hline
{Roman Forum(PR)}  & 111 & 51.1 & 10.71 & 6.75&\textbf{1.59} & \textbf{0.89} & 1.63 & 0.93 & 23.23 & 11.20 & 1.85 & 1.04\\
\hline
{Notre Dame}  & 214& 96.6 & 0.57 & 0.34 & \textbf{0.38} & \textbf{0.21} & 0.38 & 0.21 & 6.87 & 4.75 & 1.02 & 0.26\\
\hline
{Montreal N.D.}  & 162 & 97.0 &\textbf{ 0.38} & \textbf{0.24} & 0.56 & 0.37 & 0.57 & 0.38 & 10.33 & 11.15 & 0.58 & 0.41\\
\hline
{Union Square}& 144& 28.6 &5.64& 3.99 & \textbf{4.31 }& \textbf{3.76} & 4.85 & 4.38 & 9.59 & 6.69 & 5.77 & 4.83\\
\hline
{Gendarmenmarkt} & 112 & 89.7  & 45.34 & 23.63 & 37.93 & 17.35 & \textbf{37.92} & 17.41 & 62.69 & 26.42 & 54.38 & \textbf{15.91}\\
\hline
Piccadilly(PR) & 169& 55.4 & \textbf{0.73} & \textbf{0.39} & 3.68 & 1.90 & 3.71 & 1.93 & 13.55& 13.34 & - & -\\
 \hline
\end{tabular}

\end{table}

\begin{table}[htp!]
    \caption{ {Rotation errors for Photo Tourism. $N$ is the total number of cameras. $n$ is the size after down sampling. Est. \% is the ratio of observed blocks over total number of blocks. $\bar{e}_r$ is the mean rotation error, $\hat{e}_r$ is the median rotation error. The notation PR means that the dataset was further down sampled to match the two view methods. We were not able to get results for our subsampled dataset for Piccadilly with MPLS.}}
    \label{tab:Photo_Comprehensive_rotations}
    \centering
     \tabcolsep=0.055cm
\begin{tabular}{||c | c c  c c c | c c |cc| c ||} \hline
 &\multicolumn{5}{|c|}{Our Approach}  &\multicolumn{2}{|c|}{LUD} &\multicolumn{2}{|c|}{MPLS} & \\
 \hline
   dataset & N & n & Est. \% &  $\bar{e}_r$ & $\hat{e}_r$&  $\bar{e}_r$ & $\hat{e}_r$&  $\bar{e}_r$ & $\hat{e}_r$&  Our Runtime (s)\\
 \hline\hline
  {Piazza del Popolo} &307& 185 & 72.3  & 1.26 & 0.61 & 0.72 & 0.43 & \textbf{0.69}& \textbf{0.41}&13531\\
  \hline
  {NYC Library} &306& 127 & 64.7 & 2.80 & 1.58 & \textbf{1.16} & 0.61 &  1.19& \textbf{0.57}&4465 \\
  \hline
 {Ellis Island } &223& 194 & 70.3  & 4.61 & 1.11& 1.16 & 0.50 & \textbf{0.99} &\textbf{0.49}&13816 \\
 \hline
  {Tower of London} &440 & 130 & 34.1 & 2.28 & 1.31 & \textbf{1.63} & \textbf{1.28} & 1.66&1.37&4242 \\
 \hline
{Madrid Metropolis} &315& 190 & 35.9  & 28.85 & 4.60 & \textbf{1.27} & \textbf{0.61}  & 1.54&1.15&11764\\
  \hline
   {Yorkminster} & 410 &196 & 37.2 & 2.33& 1.97 & \textbf{1.34} & 1.09 & 1.89 & \textbf{1.04} & 13115\\
  \hline
{Alamo} & 564 & 224 & 94.3 & 1.10 & 0.76 & \textbf{1.07}  & \textbf{0.68} & 1.09&\textbf{0.68}&17513 \\
\hline
{Vienna Cathedral} & 770& 197& 97.8  & 0.74 & 0.46 &0.40 & \textbf{0.28} & \textbf{0.39}&\textbf{0.28}&12499 \\
\hline
{Roman Forum(PR)}  & 989& 111 & 51.1 & 11.86 & 3.39 & \textbf{0.40} & \textbf{0.28} & 1.07&0.65& 2162 \\
\hline
{Notre Dame}  & 547& 214& 96.6 & 0.78 &  0.50 & \textbf{0.67} & \textbf{0.43} & 0.68&\textbf{0.43}&17430 \\
\hline
{Montreal N.D.}  & 442& 162 & 97.0 & 0.50 & 0.35 & \textbf{0.49} & 0.32 & \textbf{0.49}&\textbf{0.31}& 7241 \\
\hline
{Union Square}& 680& 144& 28.6 & 20.70 & 5.29  & \textbf{1.82} & \textbf{1.34} & 2.00&1.56&4355 \\
\hline
{Gendarmenmarkt} &655 & 112 & 89.7  & 22.95 & 15.24 & 18.42 & 10.25 & \textbf{17.42}&\textbf{8.41}&2432\\
\hline
Piccadilly(PR)& 1000 & 169& 55.4 &\textbf{ 2.01} & \textbf{0.96} & 6.12 & 2.95 & -& - & 11230 \\
 \hline
\end{tabular}

\end{table}

\end{document}